\documentclass[10pt,twocolumn,letterpaper]{article}

\usepackage{iccv}
\usepackage{times}
\usepackage{epsfig}
\usepackage{graphicx}
\usepackage{amsmath}
\usepackage{amssymb}

\usepackage{amsfonts}
\usepackage{bbm}
\usepackage[ruled,noline]{algorithm2e}
\usepackage{algpseudocode}
\SetAlCapHSkip{0.0em}
\usepackage{multicol}
\usepackage{multirow}
\usepackage{booktabs}
\usepackage{tikz}
\usepackage{bm}
\usepackage{tabularray}
\usepackage{pifont}
\usepackage{colortbl}
\usepackage{xcolor}
\usepackage{array}
\usepackage{setspace,calc}
\usepackage{diagbox}
\usepackage{subfig}

\newcommand{\cmark}{\ding{51}}%
\newcommand{\xmark}{\ding{55}}%

\usepackage[pagebackref=true,breaklinks=true,colorlinks,bookmarks=false]{hyperref}

\iccvfinalcopy 

\def\iccvPaperID{9707} 

\ificcvfinal\fi

\newcommand\Tstrut{\rule{0pt}{2ex}}         
\newcommand\Bstrut{\rule[-1ex]{0pt}{0pt}}   

\begin{document}

\hbadness=2000000000
\vbadness=2000000000
\hfuzz=100pt

\setlength{\abovedisplayskip}{1.5pt}
\setlength{\belowdisplayskip}{1.5pt}
\setlength{\textfloatsep}{10pt plus 1.0pt minus 1.0pt}
\newcolumntype{a}{>{\columncolor{Gray}}m{1.3em}}

\setlength{\parskip}{0pt}

\title{Multi-Modal Continual Test-Time Adaptation for 3D Semantic Segmentation}

\author{Haozhi Cao, Yuecong Xu, Jianfei Yang, Pengyu Yin, Shenghai Yuan, Lihua Xie\\
School of Electrical and Electronic Engineering, Nanyang Technological University, Singapore\\
50 Nanyang Avenue, Singapore 639798\\
{\tt\small \{haozhi002, xuyu0014, yang0478, pengyuu001, syuan003\}@e.ntu.edu.sg, elhxie@ntu.edu.sg}
}

\maketitle
\ificcvfinal\thispagestyle{empty}\fi

\begin{abstract}
Continual Test-Time Adaptation (CTTA) generalizes conventional Test-Time Adaptation (TTA) by assuming that the target domain is dynamic over time rather than stationary. In this paper, we explore Multi-Modal Continual Test-Time Adaptation (MM-CTTA) as a new extension of CTTA for 3D semantic segmentation. The key to MM-CTTA is to adaptively attend to the reliable modality while avoiding catastrophic forgetting during continual domain shifts, which is out of the capability of previous TTA or CTTA methods. To fulfill this gap, we propose an MM-CTTA method called Continual Cross-Modal Adaptive Clustering (CoMAC) that addresses this task from two perspectives. On one hand, we propose an adaptive dual-stage mechanism to generate reliable cross-modal predictions by attending to the reliable modality based on the class-wise feature-centroid distance in the latent space. On the other hand, to perform test-time adaptation without catastrophic forgetting, we design class-wise momentum queues that capture confident target features for adaptation while stochastically restoring pseudo-source features to revisit source knowledge. We further introduce two new benchmarks to facilitate the exploration of MM-CTTA in the future. Our experimental results show that our method achieves state-of-the-art performance on both benchmarks.
\end{abstract}

\section{Introduction}
Test-Time Adaptation (TTA) proposes a realistic domain adaptation scenario, which adapts pre-trained models to the target domain during the testing process. Unlike previous Unsupervised Domain Adaptation (UDA) \cite{hoffman2018cycada,wu2019squeezesegv2,zou2019confidence,xu2022aligning}, TTA performs adaptation online without accessing the data from the source domain. Typical TTA \cite{wang2020tent,shin2022mm,su2022revisiting} methods assume a stationary target domain, while real-world scenarios are dynamic over time. To fulfill this gap, a previous work \cite{wang2022continual} proposes a general extension of TTA named Continual Test-Time Adaptation (CTTA) which assumes that the target domain is continually changing rather than static. Compared to TTA, CTTA is more challenging since the performance is easily affected by error accumulation \cite{chen2019progressive} and catastrophic forgetting \cite{mccloskey1989catastrophic} during the continual adaptation.

\begin{figure}[t]
\centering{
\includegraphics[width=.9\linewidth]{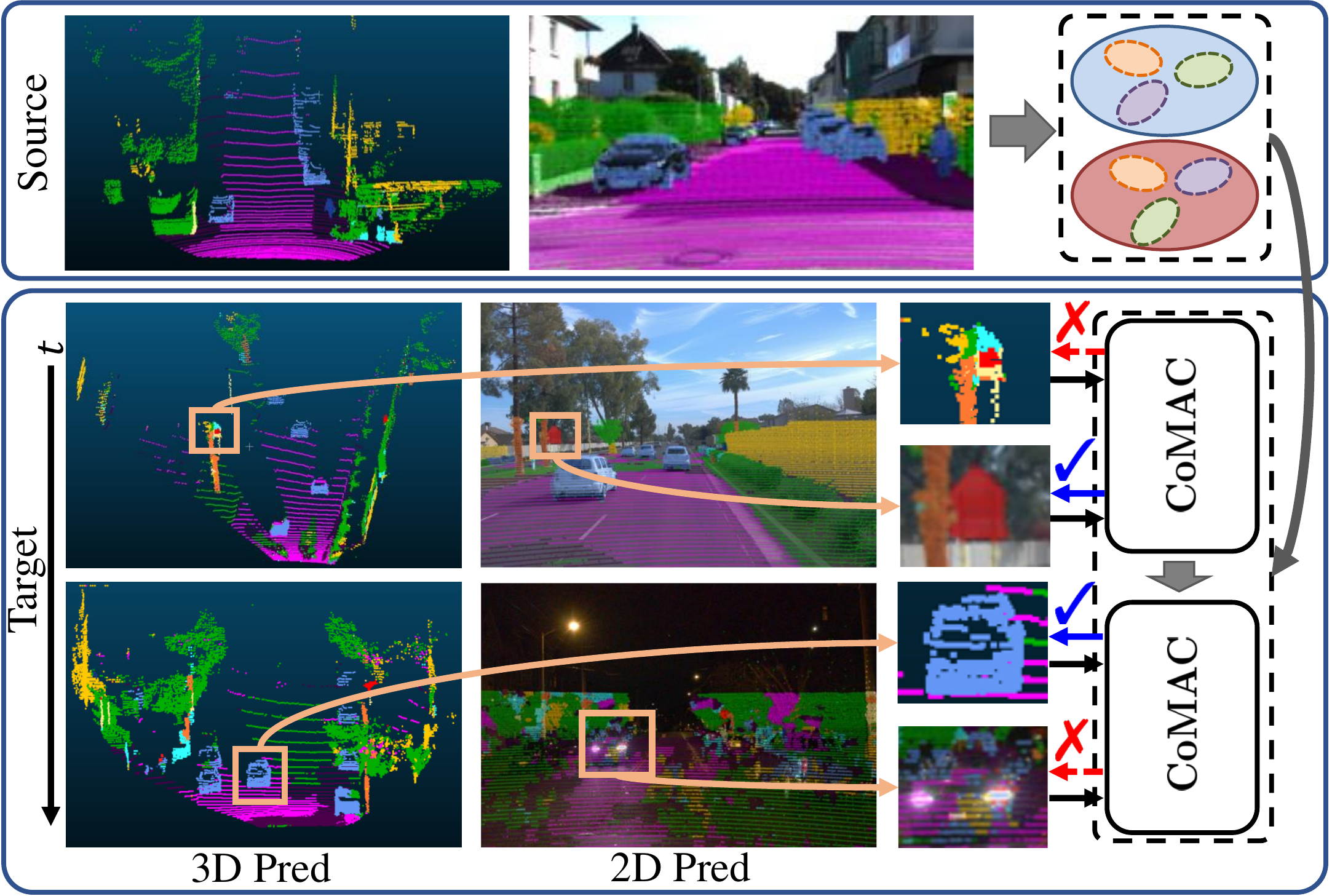}}
\caption{Illustration of how continual domain shifts affect multi-modal segmentation and our method. Unlike the source domain where predictions from both modalities are reliable, the reliability of each modality varies in MM-CTTA due to different domain shifts. CoMAC tackles MM-CTTA by attending to the reliable modality (\textcolor{blue}{\cmark}) rather than the noisy one (\textcolor{red}{\xmark}). Meanwhile, source knowledge is stochastically revisited to avoid catastrophic forgetting. Figure best viewed in color and zoomed in.}
\label{Fig:Intro}
\end{figure}

For 3D semantic segmentation, multi-modal sensors are frequently leveraged in different tasks, such as scene understanding \cite{bultmann2023real,mccormac2018fusion} and semantic map construction \cite{mccormac2017semanticfusion,wang2021multi}. For some specific applications like semantic-based localization \cite{chen2019suma++} and autonomous driving \cite{xu2021rpvnet, wu2018squeezeseg}, multi-modal information is the key to robust performance under adverse conditions. However, their collaboration has been proven to be sensitive toward domain drifts \cite{bayoudh2021survey,shin2022mm}. In real-world scenarios, such collaboration deterioration could be more severe considering that the target domain is continually changing as in CTTA (e.g., the operating environments of 24/7 AGVs are continually changing due to altering weather or illumination conditions). Hence, it is essential for multi-modal networks to adapt to the dynamic target domain in an online manner. In this work, we aim to study Multi-Modal Continual Test-Time Adaptation (MM-CTTA) for 3D semantic segmentation, where networks are continually adapted to a changing target domain taking 3D point clouds and 2D images as input without accessing the source data.

Intuitively, one can address MM-CTTA by utilizing CTTA \cite{niu2022efficient,wang2022continual} or TTA \cite{boudiaf2022parameter,sun2020test} methods on 2D and 3D networks separately. However, this simple extension can not achieve satisfactory performance since it cannot correctly attend to the reliable modality for adaptation when others suffer from severe domain shifts. Take Fig.~\ref{Fig:Intro} as an example: predictions from the 2D image are more accurate at the beginning of the target domain, while 2D results become unreliable and 3D predictions prevail as the illumination level significantly changes over time. Although previous CTTA or TTA methods have attempted to mitigate this intra-modal prediction noise by augmentations \cite{wang2022continual} or entropy minimization \cite{wang2020tent,niu2022efficient}, the domain drift in Fig.~\ref{Fig:Intro} is too severe to be effectively rectified by image input itself, leading to inevitable error accumulation. Previous works have proposed different cross-modal fusion methods to mitigate the effect of the noisy modality during adaptation, such as cross-modal consistency \cite{jaritz2020xmuda} for UDA or pseudo-label fusion based on student-teacher consistency \cite{shin2022mm} for TTA. However, their methods rely on the premise that the target domain is static, and therefore suffer from catastrophic forgetting in MM-CTTA, leading to degenerated results.


For MM-CTTA, an ideal solution is to suppress the contribution from the noisy modality and attend more to the reliable one in an online manner. Typically, the reliability of prediction can be estimated based on its corresponding feature location in the latent space. A prediction with features close to the centroid is more likely to be reliable, suffering less from the domain shift and vice versa. On the other hand, to ensure the validity of centroid-based reliability estimation, class-wise centroids should actively adapt to the continually changing target domain while stochastically revisiting the source knowledge to avoid catastrophic forgetting. To this end, we propose an MM-CTTA method called \textbf{C}ontinual Cr\textbf{o}ss-\textbf{M}odal \textbf{A}daptive \textbf{C}lustering (CoMAC) for 3D semantic segmentation. CoMAC consists of three main modules: (i) Intra-Modal Prediction Aggregation (iMPA), (ii) Inter-Modal Pseudo-Label Fusion (xMPF), and (iii) Class-Wise Momentum Queues (CMQs). On one hand, the proposed iMPA and xMPF are utilized to suppress prediction noise based on the centroid-based reliability estimation from intra-modal and inter-modal perspectives, respectively. On the other hand, CMQs are designed to actively adapt class-wise centroids for iMPA and xMPF while avoiding catastrophic forgetting. Additionally, a class-wise contrastive loss is introduced to regularize the extracted features from drifting too far from centroids.


In summary, our main contributions are three-fold. (i) We explore a new task MM-CTTA where multi-modal input is utilized to perform continual test-time adaptation for 3D semantic segmentation and propose an effective method named CoMAC to leverage multi-modal information for MM-CTTA. (ii) We propose iMPA and xMPF to generate accurate cross-modal pseudo-labels by attending to a reliable modality. CMQs are introduced to actively adapt to the target domain without catastrophic forgetting. (iii) We introduce two 3D semantic segmentation benchmarks to facilitate the future exploration of MM-CTTA. Extensive experiments show that our method achieves state-of-the-art performance, outperforming previous methods significantly by $6.9\%$ on the challenging benchmark.

\section{Related Works}
\textbf{Test-Time Adaptation}
Describing a more realistic adaptation scenario, Test-Time Adaptation (TTA) is receiving more and more attention. Different from previous Unsupervised Domain Adaptation (UDA), TTA forbids access to raw source data and adapts the source pre-trained model during test time. As one of the primary works, TENT \cite{wang2020tent} highlights the fully test-time setting and proposes to update the batch normalization layers by entropy minimization. The following works mainly address TTA by aligning batch normalization statistics \cite{niu2022efficient,lim2023ttn,zhao2023delta}, self-training with pseudo labeling \cite{goyaltest, wang2022towards}, feature alignment \cite{liu2021ttt++}, or augmentation invariance \cite{zhang2021memo,kundu2022balancing}. The aforementioned TTA methods strictly follow the one-pass protocol as mentioned in \cite{su2022revisiting}, where networks immediately infer each sample in an online manner and forbid multiple training epochs. On the other hand, some previous works \cite{liang2020we,fleuret2021uncertainty,chen2022contrastive,xu2022source} follow the multi-pass protocol by adapting the model for multiple epochs in an offline manner. As one of the primary works, Source Hypothesis Transfer (SHOT) \cite{liang2020we} proposes to update only the encoder parameters and align source and target representation by entropy minimization and pseudo-labeling. While most existing TTA methods are proposed for image classification, some recent works \cite{fleuret2021uncertainty, kundu2021generalize, liu2021source} aim to explore TTA for image semantic segmentation. Specifically, \cite{fleuret2021uncertainty} proposes to minimize the prediction entropy while maximizing its robustness toward feature noise. \cite{kundu2021generalize} divides the TTA problem into source domain generalization and target domain adaptation. \cite{liu2021source} proposes a dual attention distillation method to transfer contextual knowledge and patch-level pseudo labels for self-supervised learning.

\textbf{Continual Test-Time Adaptation}
The definition of Continual Test-Time Adaptation (CTTA) is first proposed in CoTTA\cite{wang2022continual}, which aims to adapt the model to continually changing target domains in an online manner without any source data. Specifically, CoTTA proposes to use the moving teacher model and augmentation-average predictions for noise suppression and the model stochastic restoration to avoid catastrophic forgetting. Following the scheme of CoTTA, some recent works \cite{gan2022decorate,gongnote,niu2022efficient} have addressed CTTA from different perspectives. Specifically, \cite{gongnote} leverages the temporal correlations of streamed input by reservoir sampling and instance-aware batch normalization. \cite{gan2022decorate} proposes domain-specific prompts and domain-agnostic prompts to preserve domain-specific and domain-shared knowledge, respectively. EATA \cite{niu2022efficient} performs adaptation on non-redundant samples for an efficient update.

\textbf{Multi-Modal Adaptation for Segmentation}
Thanks to the emerging multi-modal datasets \cite{behley2019semantickitti,caesar2020nuscenes,sun2020scalability,ros2016synthia,yang2022deep}, recent works start to explore how to leverage multi-modal information between 2D images and 3D point clouds to perform domain adaptation under different settings. xMUDA \cite{jaritz2020xmuda} proposes the first Multi-Modal Unsupervised Domain Adaptation (MM-UDA) method for 3D semantic segmentation. Specifically, it leverages the cross-modal prediction consistency by minimizing the prediction discrepancy from additional classifiers. Following the scheme of cross-modal learning, DsCML \cite{peng2021sparse} designs a deformable mapping between pixel-point correlation for MM-UDA while \cite{liu2021adversarial} introduces adversarial training to mitigate domain discrepancy. In addition to UDA settings, \cite{shin2022mm} proposes the first multi-modal test-time adaptation for semantic segmentation which generates intra-modal and inter-modal pseudo labels by attending to the one with more consistent predictions across student and teacher models. 

In this work, we propose MM-CTTA as a new extension of CTTA with a specific method CoMAC. While our proposed dual-stage modules (i.e., iMPA and xMPA) and CMQs share some similar merit with previous TTA methods \cite{shin2022mm,su2022revisiting}, our method is explicitly designed for MM-CTTA. Unlike previous work \cite{shin2022mm} measuring prediction reliability by teacher-student consistency, our reliability measurement relies on the feature-centroid distance to encourage feature clustering around the adapting centroids. Different from \cite{su2022revisiting} utilizing queues to measure target distribution, the objective of our CMQs is to actively update class-wise centroids to ensure the validity of iMPA and xMPA while avoiding catastrophic forgetting.

\begin{figure*}[t]
\centering{
\includegraphics[width=.83\textwidth]{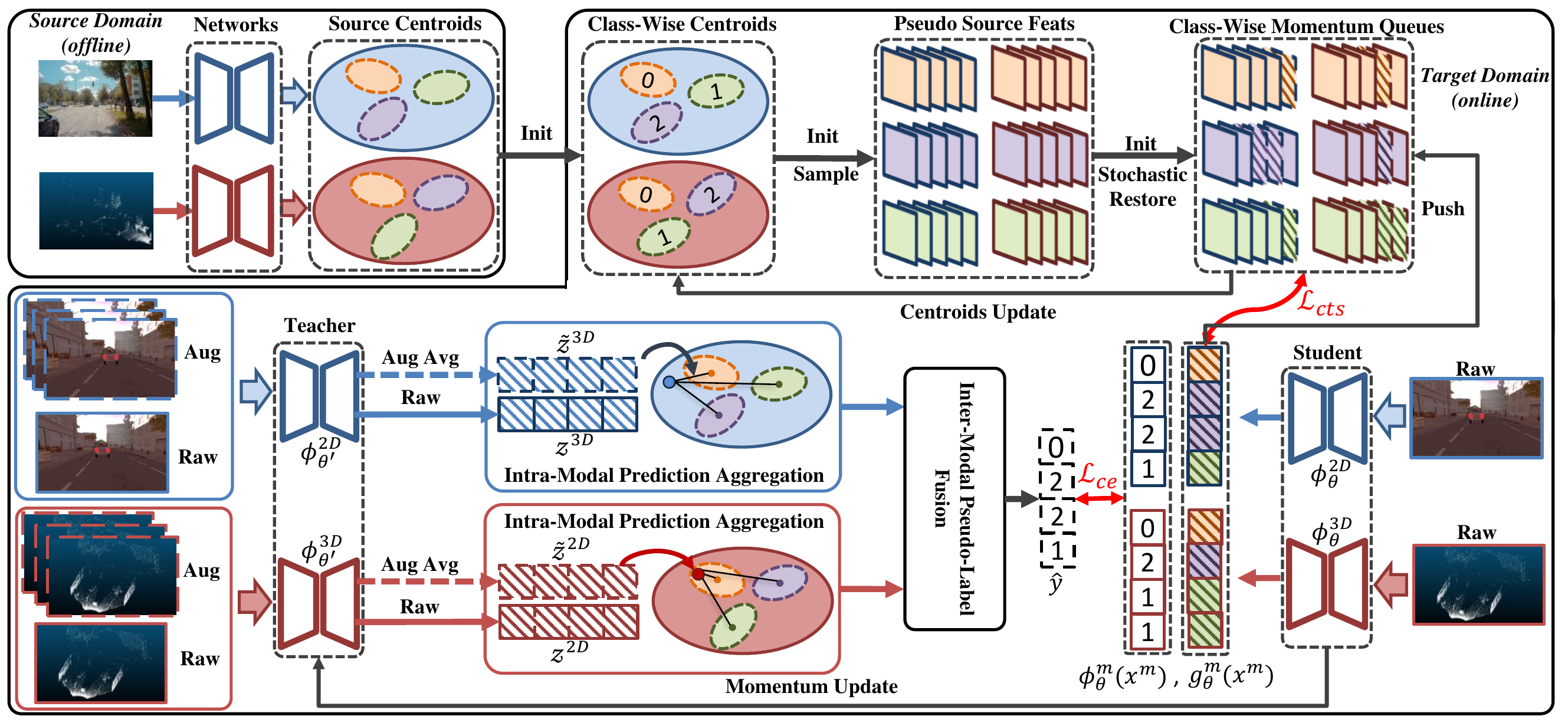}}
\caption{The main structure of our proposed method. From the source domain, we preserve the source centroids and the pre-trained model for each modality as the initialization of the class-wise centroids and all the models in the target domain, respectively. To generate reliable intra-modal predictions, Intra-Modal Prediction Aggregation (iMPA) attends to the reliable prediction whose feature shares a closer distance with the class-wise centroids in an intra-modal manner. Inter-Modal Pseudo-Label Fusion (xMPF) then fuses the intra-modal predictions by estimating the reliability of predictions from each modality for noise suppression. Class-Wise Momentum Queues (CMQs) are designed to achieve a good balance between target domain adaptation and source knowledge preservation.}
\label{Fig:Main_Method}
\vspace{-1em}
\end{figure*}

\section{Proposed Method}
\textbf{Problem definition and notations}. At timestamp $t$, the 2D image $x_{\mathcal{T},t}^{\textrm{2D}} \in \mathbb{R}^{H\times W\times 3}$ and the 3D point cloud $x_{\mathcal{T},t}^{\textrm{3D}} \in \mathbb{R}^{N\times4}$ are observed in the target domain $\mathcal{T}$, where $N$ denotes the number of 3D points located in the camera FOV. Modal-specific pre-trained networks $\phi_{\theta,t}^{m}(\cdot)=f^{m}_t(g^{m}_t(\cdot)),m\in \{\textrm{2D},\textrm{3D}\}$ consisting of the feature extractor $f^{m}_t$ and the classifier $g^{m}_t$ are used to predict the semantic labels for each point. Inspired by the fact that moving average models can provide more stable predictions \cite{tarvainen2017mean}, we utilize a fast student network $\phi_\theta^m(\cdot)=f^{m}_\theta(g^{m}_\theta(\cdot))$ and a slow teacher network $\phi_{\theta'}^m(\cdot)=f^{m}_{\theta'}(g^{m}_{\theta'}(\cdot))$ for each modality similar to previous works \cite{shin2022mm,wang2022continual}. Given $x_{\mathcal{T},t}^{m}$ as input, the features extracted by $f^{m}_t$ is denoted as $z^{m}_{\mathcal{T},t}\in\mathcal{R}^{N\times F^m}$, where $F^m$ denotes the channel number. Here we adopt the same projection protocol of previous works \cite{jaritz2020xmuda,shin2022mm} for cross-modal alignment, which projects features from the 2D branch back to 3D points, resulting in the 2D feature $z^{\textrm{2D}}_{\mathcal{T},t}$ of shape $N\times F^{\textrm{2D}}$. By default, the subscript of the target domain $\mathcal{T}$ and the timestamp $t$ are omitted for clarity. Given the multi-modal input $x_{\mathcal{T},t}^{\textrm{2D}}$, $x_{\mathcal{T},t}^{\textrm{3D}}$, the goal of MM-CTTA is to output reliable cross-modal predictions by continuously adapting to the changing target domain. In this work, we interpret the core of MM-CTTA as two-fold: (i) attending to the reliable modality for noisy suppression, and (ii) revisiting source knowledge to prevent catastrophic forgetting.

To this end, we propose Continual Cross-Modal Adaptive Clustering (CoMAC) to tackle MM-CTTA from the aforementioned perspectives. Specifically, as shown in Fig.~\ref{Fig:Main_Method}, the class-wise centroids are initialized from the source domain and pseudo-source features are randomly sampled around the centroids as source knowledge (Sec.~\ref{Section: Method-Init}). Given the raw and augmentation-average predictions from the teacher models as input, Intra-Modal Prediction Aggregation (iMPA) generates stable intra-modal predictions as their weighted sum based on their feature distance to the class-wise centroids (Sec.~\ref{Section: Method-iMPA}). Inter-Modal Pseudo-Label Fusion (xMPF) then combines the intra-modal predictions from each modality based on their reliability and output cross-modal pseudo-labels as supervision signals for student networks (Sec.~\ref{Section: Method-xMPF}). Given raw samples as input, confident target features from student networks are utilized to update Class-Wise Momentum Queues (CMQs), while pseudo-source features are stochastically restored to avoid catastrophic forgetting (Sec.~\ref{Section: Method-CMQ}). 

\subsection{Source Models and Class-Wise Centroids}\label{Section: Method-Init}
To effectively avoid catastrophic forgetting and inspired by previous TTA methods \cite{EasMasWilSch21,su2022revisiting} which preserve trivial source domain information, we utilize pseudo-source features sampled around source offline feature centroids as source representatives. The source offline centroids and pseudo-source features are treated as the prior knowledge from the source domain, which plays an essential role in preventing catastrophic forgetting detailed in Sec.~\ref{Section: Method-CMQ}. Specifically, given a specific semantic category $k$, the corresponding source offline centroid is modeled as Gaussian distribution, and pseudo-source features are denoted as:
\begingroup
\allowdisplaybreaks
\begin{align}
C_{src}^{m,k} &= \mathcal{N}(\mu_{src}^{m,k}, \sigma_{src}^{m,k}),\label{Equation:offline_centroids}\\
\mathcal{Z}^{m,k}_{src} &= \{z^{m,k}_{src,i}\sim C_{src}^{m,k}\mid i\in [1,N_q]\},\label{Equation:Pseudo_Source}
\end{align}
\endgroup
where $N_q$ is the number of pseudo-source features. $\mu_{src}^{m,k}$ and $\sigma_{src}^{m,k}$
denote the mean and standard deviation of normalized source features from the category $k$, respectively. Here we use the features generated by the pre-trained extractor $g_0^m(\cdot)$ for the centroid construction.

\subsection{Intra-Modal Prediction Aggregation}\label{Section: Method-iMPA}
In MM-CTTA, the prediction from each modality could be unreliable due to continual domain shifts, causing severe noise to the multi-modal fusion. To mitigate the intra-modal noise, iMPA aims to generate reliable intra-modal predictions for each modality. Although previous methods \cite{wang2022continual, sun2020test} regard the augmentation-average prediction as a more stable alternative, we argue that its superiority is not certain since test-time augmentations may introduce inductive bias to the prediction \cite{shanmugam2021better} (e.g., resizing image as in \cite{wang2022continual} could cause ambiguity to small-scale classes). Unlike the previous work \cite{wang2022continual}, we propose an adaptive mechanism to fuse the raw and the augmentation-average prediction as the weighted sum based on their feature distance between class-wise centroids. Given the input sample $x^m$ and its augmented variants $\{\widetilde{x}^m_i\mid i\in[1,N^m_{aug}]\}$, the features and predictions are extracted by the teacher model $\phi_{\theta'}^m$ as:
\begingroup
\allowdisplaybreaks
\begin{align}
p(x^m) &= f^m_{\theta'}(g^m_{\theta'}(x^m)) = f^m_{\theta'}(z^m),\label{Equation:Teacher_Inf_Start}\\
\widetilde{p}(x^m) &= \frac{1}{N^m_{aug}}\sum_{i=1}^{N^m_{aug}}f^m_{\theta'}(g^m_{\theta'}(\widetilde{x}^m_i)),\label{Equation:Teacher_Inf_End}
\end{align}
\endgroup
where $p(x^m)$ and $\widetilde{p}(x^m)$ represent the raw and the augmentation-average prediction, respectively. Both predictions are of shape $N\times N_c$, where $N_C$ is the number of classes and $N^m_{aug}$ denotes the number of augmentations. The normalized augmentation-average feature is computed as $\widetilde{z}^m=\sum_{i=1}^{N^m_{aug}}g^m(\widetilde{x}^m_i)/(\|g^m(\widetilde{x}^m_i)\|\cdot N^m_{aug})$. Subsequently, the weights for predictions $p(x^m), \widetilde{p}(x^m)$ are computed based on their corresponding feature distance to the class-wise centroid in a point-wise manner, so that each point of the intra-modal prediction $\hat{p}(x^m)$ can attend more to the confident prediction located closer to the cluster in the latent space. Taking a point $(j)$ as an example, given its predictions $p_{(j)}(x^{m})$, $\widetilde{p}_{(j)}(x^{m})$ and features $z^{m}_{(j)}$, $\widetilde{z}^{m}_{(j)}$:
\begingroup
\allowdisplaybreaks
\begin{align}
w^{m}_{(j)} &= \frac{\exp(\langle \mu^{m,k}, z^{m}_{(j)}\rangle)}{\sum_{i\in N_c}\exp(\langle \mu^{m,i},z^{m}_{(j)}\rangle)},\label{Equations:iMPA-start} \\
\widetilde{w}^{m}_{(j)} &= \frac{\exp(\langle \mu^{m,\widetilde{k}},\widetilde{z}^{m}_{(j)} \rangle)}{\sum_{i\in N_c}\exp(\langle \mu^{m,i},\widetilde{z}^{m}_{(j)} \rangle)},\\
\hat{p}_{(j)}(x^m) &= \frac{w^{m}_{(j)} p_{(j)(x^m)} + \widetilde{w}^{m}_{(j)} \widetilde{p}_{(j)}(x^m)}{w^{m}_{(j)} + \widetilde{w}^{m}_{(j)}},\label{Equations:iMPA-end}
\end{align}
\endgroup
where $w^{m}_{(j)}$, $\widetilde{w}^{m}_{(j)}$ are termed as the raw weight and augmentation-average weight, respectively. $k$, $\widetilde{k}$ are the largest element locations of $p_{(j)}(x^m)$, $\widetilde{p}_{(j)}(x^m)$. $\mu^{m,k}$ denotes the mean of class $k$ centroid, which is actively updated as shown in Sec.~\ref{Section: Method-CMQ}. $\langle a,b \rangle$ represents the inner product of vectors $a$ and $b$. Our empirical results show that this adaptive fusion between the raw and augmentation-average predictions results in more stable intra-modal predictions.

\subsection{Inter-Modal Pseudo-Label Fusion}\label{Section: Method-xMPF}
Given the output of iMPA, the goal of xMPF is to generate pseudo-labels by estimating their reliability in a cross-modal manner. Since domain shifts can affect the reliability of each modality variously (e.g., images in day and night), the cross-modal pseudo-label should adaptively attend to the reliable modality for noise suppression. Motivated by this idea, the proposed xMPF generates the cross-modal pseudo-labels as the weighted sum of intra-modal predictions from iMPA. Specifically, taking point $(j)$ as an example, the inter-modal weight $\hat{w}^m_{(j)}$ is computed as the summation of both weights if their corresponding predictions indicate the same class, or the maximum one otherwise. This can be viewed as an implicit way to encourage point-wise prediction consistency for each modality. Formally, the inter-modal weight $\hat{w}^m_{(j)}$ is computed as follows:
\begin{equation}\label{Equation:Weight_xMPF}
\hat{w}^m_{(j)}=\begin{cases}
w^{m}_{(j)} + \widetilde{w}^{m}_{(j)},\ \ \textrm{if} \ \ k=\widetilde{k}\\
\max(w^{m}_{(j)}, \widetilde{w}^{m}_{(j)}),\ \ \textrm{if} \ \ k\neq\widetilde{k}
\end{cases}.
\end{equation}
The cross-modal prediction is then computed as the weighted sum in a cross-modal manner:
\begingroup
\allowdisplaybreaks
\begin{align}
\hat{p}^{\textrm{xM}}_{(j)}(x) &= \frac{\hat{w}^{\textrm{2D}}_{(j)} \hat{p}^{\textrm{2D}}_{(j)}(x) + \hat{w}^{\textrm{3D}}_{(j)} \hat{p}^{\textrm{3D}}_{(j)}(x)}{\hat{w}^{\textrm{2D}}_{(j)} + \hat{w}^{\textrm{3D}}_{(j)}},\\
\hat{y}_{(j)} &= \arg \max_{k\in N_c}(\hat{p}^{\textrm{xM}}_{(j)}(x)^{(k)}),\label{Equation:xMPsLabel}
\end{align}
\endgroup
where $\hat{p}^{\textrm{xM}}_{(j)}(x)$, $\hat{y}_{(j)}$ denote the cross-modal prediction and pseudo-label, respectively. The cross-modal pseudo-label can therefore attend to the more reliable modality with confident intra-modal predictions.

\subsection{Class-Wise Moving Queues}\label{Section: Method-CMQ}
Both iMPA and xMPF greatly depend on the quality of class-wise centroids. Initialized from the source offline centroids as in Eq.~\ref{Equation:offline_centroids}, the class-wise centroids should continually adapt to the target domain to ensure the validity of iMPA and xMPF. Meanwhile, the source knowledge should be occasionally played back during the centroid adaptation so that it can be revisited during the reliability estimation in iMPA and xMPF without catastrophic forgetting. To achieve a balance between adaptation and source knowledge preservation, we propose CMQs which utilize momentum queues \cite{he2020momentum} to actively estimate feature clustering of the target domain by capturing confident target features while stochastically restoring pseudo-source features for each modality. Specifically, the CMQ of the class $k$ is initialized by the pseudo-source features $\mathcal{Z}^{m,k}_{src}$ as in Eq.~\ref{Equation:Pseudo_Source}. For each iteration, taking the raw sample as input, the normalized point features from the student network $\phi_\theta^m$ with confident predictions are preserved by threshold-based filtering. Given the CMQ of class $k$ from the previous step denoted as $\mathcal{\hat{Q}}_{t-1}^{m,k}$, the CMQ of current step is updated as:
\begingroup
\allowdisplaybreaks
\begin{align}
\mathcal{Z}_{cf}^{m,k} &= \mathbbm{1}^k_{cf}[g_\theta^m(x^m)/\|g_\theta^m(x^m)\|],\label{Equation:CMQ_Start} \\
\mathcal{\widetilde{Q}}_{t}^{m,k} &= \Phi(\mathcal{\hat{Q}}_{t-1}^{m,k}, \mathcal{Z}_{cf}^{m,k}),\\
\mathcal{\hat{Q}}_{t}^{m,k} &= \begin{cases}
\mathcal{\widetilde{Q}}_{t}^{m,k}, &\textrm{if} \ \ \gamma_t> p_{rs}\\
\Phi(\mathcal{\hat{Q}}_{t-1}^{m,k}, \mathbbm{1}^{k}_{rs}\mathcal{Z}^{m,k}_{src}), &\textrm{if} \ \ \gamma_t\leq p_{rs}
\end{cases},
\end{align}
\endgroup
where $\mathbbm{1}^k_{cf}$ is the confidence index for class $k$ where $\mathbbm{1}^k_{cf}\phi_\theta^m(x^m)^{(k)}\geq\tau_{cf}$ and $\mathbbm{1}^{k}_{rs}$ denotes the source restoring index which randomly samples $N_{enq}$ pseudo-source features from $\mathcal{Z}^{m,k}_{src}$. $\Phi(a,b)$ denotes the operation of enqueuing $b$ in $a$ as in \cite{he2020momentum}. $\gamma_t$ is a restoring flag uniformly sampled from $[0,1]$ at timestamp $t$ and $p_{rs}$ denotes the hyper-parameter of restoring probability. Different from the image classification problem which can capture all confident samples \cite{su2022revisiting}, the number of confident points from each class can be huge (i.e., more than 1,000), which is too aggressive and expensive to enqueue all points. Therefore, we limit the index number of $\mathbbm{1}^{k}_{cf}$ by an upper bound $N_{enq}$ through selecting features with the top $N_{enq}$ confident predictions.

For each optimization step, the mean of the class-wise centroid for the next step $\mu_{t+1}^{m,k}$ is updated as the average of CMQ of the current timestamp $\mathcal{\hat{Q}}_{t}^{m,k}$ so that the class-wise centroids can adapt to the feature clustering of target domain without catastrophic forgetting. Additionally, we propose a class-wise contrastive loss modified from \cite{khosla2020supervised} as a regularizer so that the target confident features can revisit source knowledge through clustering around the class-wise centroids. Specifically, given any confident target feature from Eq.~\ref{Equation:CMQ_Start} as the anchor, the positive samples are the features from the CMQ that shares the same modality and semantic class, denoted as $P(k)=\{\hat{q}_p \mid \hat{q}_p\in \mathcal{\hat{Q}}_t^{m,k}\}$. The negative samples are the features from the CMQs of the same modality but different classes denoted as $A(k)=\{\hat{q}_n \mid \hat{q}_n\in \mathcal{\hat{Q}}_t^{m,a}, \forall a\neq k \bigcap a\in N_c\}$. The class-wise contrastive loss $\mathcal{L}^m_{cts}$ is computed as:
\begingroup
\begin{align}
\mathcal{L}_{cts}^m = &\sum_{i\in |\mathcal{Z}_{cf}^{m,k}|} \frac{-1}{|P(k)|}\nonumber \\
&\sum_{q_p\in P(k)}\frac{\exp{(\hat{z}_{t,(i)}^{m,k} \cdot q_p)}}{\sum_{q_n\in N(k)}\exp{(\hat{z}_{t,(i)}^{m,k} \cdot q_n)}}.\label{Equation:CMQ_End}
\end{align}
\endgroup

\begin{algorithm}[t]
\SetKwInput{kwModelInit}{Init Model}
\SetKwInput{kwCMQInit}{Init CMQ}
\DontPrintSemicolon
\SetInd{0.2em}{0.2em}
\kwModelInit{Teacher $\phi_{\theta'}^m$, student $\phi_{\theta}^m$, $m\in\{\text{2D},\text{3D}\}$} \kwCMQInit{$\mathcal{\hat{Q}}_{0}^{m,k}=\mathcal{Z}^{m,k}_{src}$, $m\in\{\text{2D},\text{3D}\}, k\in N_c$}
\For{$t\in|\mathcal{X}_{\mathcal{T}}|$, $x=(x^{\text{2D}},x^{\text{3D}})\in\mathcal{X}_{\mathcal{T}}$}{
\For{$m\in\{\textrm{2D},\textrm{3D}\}$}{
1. Augmented input to get $\{\widetilde{x}^m_{i} \mid i\in N^m_{aug}\}$\;
2. Get predictions and features with $\phi_{\theta'}^m$ by Eq.~\ref{Equation:Teacher_Inf_Start}-\ref{Equation:Teacher_Inf_End}\;
3. Get intra-modal $\hat{p}^m(x^m)$ by iMPA through Eq.~\ref{Equations:iMPA-start}-\ref{Equations:iMPA-end}\;
}
4. Get inter-modal $\hat{y}$ by xMPF through Eq.~\ref{Equation:Weight_xMPF}-\ref{Equation:xMPsLabel}\;
5. Update CMQs and compute $\mathcal{L}^m_{cts}$ by Eq.~\ref{Equation:CMQ_Start}-\ref{Equation:CMQ_End}\;
6. Update $\phi_{\theta}^m$ by Eq.~\ref{Equation:Total_Loss} and $\phi_{\theta'}^m$ by Eq.~\ref{Equation:Teacher_Update}\;
}
\caption{The proposed CoMAC}\label{Algorithm:PsCode}
\end{algorithm}

\subsection{Main Structure and Optimization}\label{Section: Method-Overview}
The main structure of our proposed method is shown in Fig.~\ref{Fig:Main_Method}. Given the multi-modal input $x^m$, the teacher model is used to infer $x^m$ and its augmented variants $\{\widetilde{x}^m_i | i\in[1,N^m_{aug}]\}$ and then generate reliable cross-modal pseudo-labels denoted as $\hat{y}$ through iMPA and xMPF. The student model is directly updated by minimizing the weighted sum of the standard cross-entropy loss between its prediction and the pseudo-label $\hat{y}$, and the contrastive loss in Eq.~\ref{Equation:CMQ_End}. The teacher model is updated as the exponential moving average of the student's weights as in previous works \cite{shin2022mm,wang2022continual}:
\begingroup
\allowdisplaybreaks
\begin{align}\label{Equation:Student_Training}
\mathcal{L}^m_{ce} &= \mathbbm{1}_{\hat{y}}\log{\phi^m(x^{m})},\\
\mathcal{L}_{total} &= \sum_{m}^{\{\textrm{2D},\textrm{3D}\}}(\mathcal{L}^m_{ce} + \lambda_{cts}\mathcal{L}^m_{cts}),\label{Equation:Total_Loss}\\
\theta'_{t+1} &= (1-\lambda_s)\theta + \lambda_s\theta',\label{Equation:Teacher_Update}
\end{align}
\endgroup
where $\lambda_{cts}$ and $\lambda_s$ are the hyper-parameters denoting the coefficient of the contrastive loss and the momentum factor. Our adaptation process is summarized as Algorithm~\ref{Algorithm:PsCode}.

\section{Experiments}
In this section, we present our experimental results based on two new benchmarks. The details of our proposed benchmarks, baselines, and implementation are first introduced in Sec.~\ref{Section: Exp_Settins}. Sec.~\ref{Section: Exp_Main_Results} presents our overall results and Sec.~\ref{Section: Exp_Ablation} justifies our method by extensive ablation studies.

\begin{table*}[t]
\centering
\setlength{\tabcolsep}{4pt}
\resizebox{.97\textwidth}{!}{
	\begin{tabular}{ m{8.0em} |  m{4.7em} |
			m{1.8em}<{\centering} m{1.8em}<{\centering} m{1.8em}<{\centering} m{1.8em}<{\centering} m{1.8em}<{\centering} m{1.8em}<{\centering} m{1.8em}<{\centering} m{1.8em}<{\centering} m{1.8em}<{\centering} m{1.8em}<{\centering} m{1.8em}<{\centering} m{1.8em}<{\centering} | m{1.8em}<{\centering} | m{1.8em}<{\centering} m{1.8em}<{\centering} m{1.8em}<{\centering} m{1.8em}<{\centering} m{1.8em}<{\centering} m{1.8em}<{\centering} m{1.8em}<{\centering} m{1.8em}<{\centering} m{1.8em}<{\centering} m{1.8em}<{\centering} | m{1.8em}<{\centering}}
		\hline
		\multicolumn{2}{c|}{Time} & \multicolumn{13}{c|}{$t \ \ \xrightarrow{\hspace*{28em}}$} &\multicolumn{11}{c}{$t \ \ \xrightarrow{\hspace*{22em}}$}\\
		\hline
		\multirow{2}{*}[-2.2em]{Methods} & \multirow{2}{*}[-2.2em]{DA Type} & 
		\multicolumn{13}{c|}{SemanticKITTI-to-Synthia} & \multicolumn{11}{c}{SemanticKITTI-to-Waymo} \\
		\cline{3-26}
		& & \rotatebox{75}{01-Spring} & \rotatebox{75}{02-Summer} & \rotatebox{75}{04-Fall} & \rotatebox{75}{05-Winter} & \rotatebox{75}{01-Dawn} & \rotatebox{75}{02-Night} & \rotatebox{75}{04-Sunset} & \rotatebox{75}{05-W-night} & \rotatebox{75}{01-Fog} & \rotatebox{75}{02-S-Rain} & \rotatebox{75}{04-R-Night} & \rotatebox{75}{05-Rain} & \cellcolor{gray!20}Avg & \rotatebox{75}{D-O-1} & \rotatebox{75}{D-P-1} & \rotatebox{75}{D-S-1} & \rotatebox{75}{DD-1} & \rotatebox{75}{N-1} & \rotatebox{75}{D-O-2} & \rotatebox{75}{D-P-2} & \rotatebox{75}{D-S-2} & \rotatebox{75}{DD-2} & \rotatebox{75}{N-2} & \cellcolor{gray!20}Avg\\
		\hline
		Source & - & 
		28.3 & 30.7 & 37.7 & 27.3 & 23.1 & 27.3 & 38.2 & 26.5 & 27.0 & 21.6 & 27.8 & 16.4 & \cellcolor{gray!20}27.7 & 33.2 & 36.9 & 29.5 & 28.3 & 14.0 & 32.8 & 38.0 & 30.4 & 28.4 & 14.6 & \cellcolor{gray!20}28.6\\
		\hline
		Pslabel & TTA & 
		29.6 & 27.1 & 36.4 & 28.4 & 21.4 & 26.6 & 32.3 & 26.1 & 27.4 & 21.8 & 25.0 & 18.0 & \cellcolor{gray!20}26.7 & 39.1 & 41.9 & 37.0 & 36.8 & 25.7 & 38.5 & 43.7 & 37.6 & 37.6 & 24.6  & \cellcolor{gray!20}36.9\\
		TENT \cite{hu2021fully} & TTA &  
		\textbf{35.4} & 24.9 & 24.9 & 20.7 & 18.7 & 16.6 & 16.0 & 13.6 & 13.6 & 11.5 & 8.4 & 9.1 & \cellcolor{gray!20}17.8 & 33.6 & 22.3 & 18.8 & 16.9 & 10.8 & 15.7 & 12.9 & 9.8 & 11.3 & 10.6 & \cellcolor{gray!20}16.3\\
		LAME \cite{boudiaf2022parameter}& TTA & 
		14.0 & 12.4 & 17.3 & 13.0 & 12.6 & 11.6 & 17.0 & 13.2 & 19.2 & 7.7 & 7.8 & 6.0 & \cellcolor{gray!20}12.7 & 11.9 & 10.4 & 13.4 & 9.2 & 8.7 & 13.4 & 11.2 & 12.2 & 9.7 & 8.6 & \cellcolor{gray!20}12.7 \\
		xMUDA-pl \cite{jaritz2020xmuda} & MM-TTA & 
		28.8 & 26.9 & 35.9 & 28.2 & 21.3 & 26.5 & 32.2 & 26.0 & 27.0 & 21.6 & 24.9 & 17.9 & \cellcolor{gray!20}26.4 & 39.3 & 42.1 & 37.4 & 37.0 & 25.9 & 38.7 & \textbf{43.9} & 37.9 & 37.7 & 25.0 & \cellcolor{gray!20}36.5\\
		MMTTA \cite{shin2022mm} & MM-TTA & 
		31.1 & 24.4 & 30.7 & 25.8 & 28.3 & 24.2 & 26.8 & 23.2 & 29.6 & 20.7 & 22.1 & 20.6 & \cellcolor{gray!20}25.6 & 39.9 & 40.0 & 30.9 & 31.5 & 29.6 & 30.6 & 32.2 & 23.9 & 26.4 & 23.8 & \cellcolor{gray!20}30.9\\
		CoTTA \cite{wang2022continual} & CTTA & 
		29.7 & 27.2 & 34.7 & 27.0 & 26.4 & 25.6 & 33.0 & 27.3 & 28.1 & 18.0 & 22.7 & 18.4 & \cellcolor{gray!20}26.5 & 32.9 & 28.0 & 22.9 & 22.2 & 20.3 & 26.1 & 24.9 & 23.6 & 24.7 & 19.7 & \cellcolor{gray!20}24.5\\
		EATA \cite{niu2022efficient} & CTTA & 
		34.0 & 30.0 & 38.6 & 30.2 & 30.2 & 28.4 & 36.5 & 30.1 & 32.2 & 21.3 & 25.3 & 20.1 & \cellcolor{gray!20}29.7 & 40.1 & 40.8 & 36.3 & 34.3 & 28.9 & 39.3 & 41.7 & 36.5 & 37.9 & 28.8 & \cellcolor{gray!20}36.5\\
		\hline
		MMTTA-rs \cite{shin2022mm} & MM-CTTA & 
		31.9 & 28.1 & 37.7 & 29.3 & 28.7 & 27.6 & 35.2 & 29.7 & 30.2 & 20.4 & 25.6 & 20.3 & \cellcolor{gray!20}28.7 & 40.4 & 41.5 & 36.3 & 34.7 & 30.2 & 39.9 & 41.5 & 36.4 & 38.3 & 29.9 & \cellcolor{gray!20}36.9 \\
		CoMAC (Ours) & MM-CTTA & 
		34.0 & \textbf{34.7} & \textbf{44.9} & \textbf{38.1} & \textbf{35.0} & \textbf{37.8} & \textbf{45.1} & \textbf{39.0} & \textbf{39.1} & \textbf{29.4} & \textbf{35.1} & \textbf{27.1} & \cellcolor{gray!20}\textbf{36.6} & \textbf{41.4} & \textbf{43.7} & \textbf{38.8} & \textbf{37.8} & \textbf{32.2} & \textbf{42.2} & 41.8 & \textbf{40.7} & \textbf{38.9} & \textbf{30.4} & \cellcolor{gray!20}\textbf{38.8}\\
		\hline
		\multicolumn{2}{c|}{Time} & \multicolumn{13}{c|}{$t \ \ \xleftarrow{\hspace*{28em}}$} &\multicolumn{11}{c}{$t \ \ \xleftarrow{\hspace*{22em}}$}\\
		\hline
		TENT \cite{hu2021fully} & TTA & 10.2 & 8.8 & 8.6 & 9.0 & 10.2 & 9.5 & 9.8 & 11.4 & 18.3 & 13.2 & 16.9 & 18.9 & \cellcolor{gray!20}12.1 & 12.0 & 11.5 & 9.2 & 9.8 & 9.7 & 12.6 & 13.0 & 12.5 & 19.8 & 21.2 & \cellcolor{gray!20}13.1\\
		MMTTA-rs \cite{shin2022mm} & MM-CTTA & 31.4 & 27.4 & 37.2 & 28.7 & 26.2 & 27.2 & 33.1 & 29.0 & 28.5 & 19.9 & 22.3 & 19.7 & \cellcolor{gray!20}27.6 & 40.4 & 41.4 & 36.4 & 31.6 & 29.1 & 37.6 & 40.2 & 35.4 & 35.2 & 27.8 & \cellcolor{gray!20}35.5\\
		CoMAC (Ours) & MM-CTTA & \textbf{34.6} & \textbf{32.8} & \textbf{40.9} & \textbf{33.2} & \textbf{31.8} & \textbf{32.4} & \textbf{39.2} & \textbf{33.2} & \textbf{33.8} & \textbf{25.1} & \textbf{30.3} & \textbf{23.3} & \cellcolor{gray!20}\textbf{32.4} & \textbf{40.8} & \textbf{41.9} & \textbf{37.7} & \textbf{33.2} & \textbf{30.3} & \textbf{39.6} & \textbf{43.6} & \textbf{38.8} & \textbf{38.6} & \textbf{30.0} & \cellcolor{gray!20}\textbf{37.5}\\
		\hline
		
	\end{tabular}
}
\smallskip
\caption{Performance (mIoU) of SemanticKITTI-to-Synthia. Here we report the softmax-average mIoU of 2D and 3D prediction.}
\label{Table:C2MAC_results}
\vspace{-4pt}
\end{table*}

\begin{figure*}[t]
\centering
\includegraphics[width=.95\textwidth]{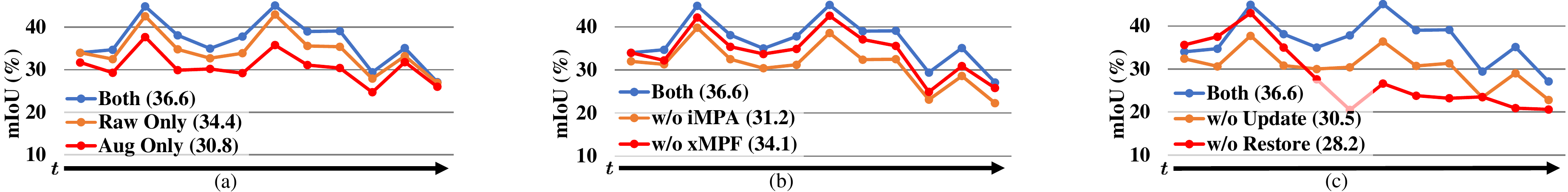}
\caption{Ablation studies to justify our design of CoMAC. Specifically, (a) compares CoMAC with variants only using raw or augmented samples as input of iMPA and (b) justifies the necessity of weighting mechanisms in iMPA and xMPF. (c) compares CoMAC with variants without either target feature enqueuing or pseudo-source features restoring in CMQs. Scores within the parentheses are the average mIoU.}
\label{Fig:Exp_Ablation}
\vspace{-8pt}
\end{figure*}

\subsection{Benchmarks and Settings}\label{Section: Exp_Settins}
\noindent \textbf{Proposed benchmarks.} To evaluate the performance under MM-CTTA settings, we proposed two benchmarks in this work, including (i) \textbf{SemanticKITTI-to-Synthia (S-to-S)} and (ii) \textbf{SemanticKITTI-to-Waymo (S-to-W)}. Both benchmarks leverage SemanticKITTI \cite{behley2019semantickitti} as the source-domain dataset, which utilizes a 0.7MP camera and a 64-line LiDAR. For the target domain, we utilize two different datasets including Synthia \cite{ros2016synthia} and Waymo \cite{sun2020scalability} thanks to the various environmental conditions they cover. The former shares a larger domain gap with more diverse seasons, weather, and illumination conditions while the latter is less challenging but closer to the real-world application. For S-to-S, it is constructed by a sequence of different videos from Synthia \cite{ros2016synthia} without any repetition. Since Synthia includes depth images instead of 3D point clouds, we generate a simulated point cloud for each sample by randomly sampling depth pixels as 3D points following the structural characteristic of LiDAR. For each sequence, all samples are included for adaptation and we use ``No.-Conditions'' to indicate their corresponding video sequence in Synthia (i.e, ``01-Spring'' indicates video sequence 01 under Spring condition). Different from the previous CTTA benchmark \cite{wang2022continual} which only includes 4 different unique sequences, the proposed S-to-S consists of 12 different sequences recorded under various environmental conditions without repetition. By default, we organize the sequences following the order of seasons$\rightarrow$illumination$\rightarrow$weathers as a challenge ascending order. For S-to-W, we sort out the illumination conditions of each sequence in Waymo dataset \cite{sun2020scalability} based on the sequence descriptions and divide all samples into three types, including Day (\textbf{D}), Dawn-Dusk (\textbf{DD}), and Night (\textbf{N}). Since the number of sequences of \textbf{D} is much larger than others, we further split the sequence of \textbf{D} based on their recording location, including Phoenix (\textbf{P}), San Francisco (\textbf{S}), and Others (\textbf{O}). Each sequence is separated in two halves to simulate the revisiting situation as in \cite{wang2022continual} but without repeated samples (i.e., ``D-O-1'' indicates the first half of the D-O sequence). Similar to S-to-S, S-to-W also follows a challenge ascending order. For both datasets, we adopt the images of the front-view camera similar to the source-domain dataset. We utilize a similar class mapping strategy as \cite{jaritz2020xmuda,shin2022mm} with slight modification since some classes are missing in the target domain. \textit{More details are illustrated in the appendix.}

\noindent \textbf{Baselines.} Previous TTA, MM-TTA, and CTTA methods are evaluated on both S-to-S and S-to-M as baselines. For TTA methods, we compare with pseudo-labels with threshold filtering (Pslabel), TENT \cite{hu2021fully} and LAME \cite{boudiaf2022parameter}, while xMUDA with pseudo-labels (xMUDA-pl) \cite{jaritz2020xmuda} and MMTTA \cite{wang2021multi} are regarded as the representatives of MM-TTA methods. CoTTA \cite{wang2022continual} and EATA \cite{niu2022efficient} are the recent works that focus on CTTA problems. Specifically, xMUDA-pl is originally designed for UDA, whereas we modify it into a TTA version by discarding the source data during adaptation. Considering our CoMAC is the only MM-CTTA method, we implement an MM-CTTA version of MMTTA \cite{shin2022mm} integrated with model-based stochastic restoration in \cite{wang2022continual} for a fair comparison. We report the softmax-average mIoU of 2D and 3D predictions as results.

\noindent \textbf{Implementation details.} Following previous works \cite{jaritz2020xmuda,shin2022mm}, we utilize UNet \cite{ronneberger2015u} with ResNet-34 \cite{he2016deep} encoder as the 2D backbone and SCN \cite{graham20183d} based on UNet as the 3D backbone for all baselines. The pre-training procedures on SemanticKITTI dataset follow \cite{jaritz2020xmuda}. For our method, we empirically choose resizing and z-rotation as the 2D and 3D augmentations following previous works \cite{wang2022continual,jaritz2020xmuda}. Specifically, we utilize multiple resizing factors of [0.5, 0.625, 0.75, 0.875] and z-rotation of [$60^{\circ}$, $120^{\circ}$, $180^{\circ}$, $240^{\circ}$, $300^{\circ}$]. $N_q$ and $N_{enq}$ are set to 4096 and 200, respectively. We utilize a restore rate $p_{rs}$ of 0.5 and a momentum factor $\lambda_s$ of 0.999 with a coefficient $\lambda_{cts}$ of 1. All methods strictly follow the one-pass protocol \cite{su2022revisiting} (i.e., the training epoch is one where inference is conducted immediately for each sample). \textit{More details are included in the appendix.}

\subsection{Overall Results}\label{Section: Exp_Main_Results}
As shown in Table~\ref{Table:C2MAC_results}, our CoMAC achieves SOTA performance on the average mIoU of all sequences for both benchmarks. Specifically, the proposed CoMAC significantly outperforms previous methods by more than $6.9\%$ on S-to-S and $1.9\%$ on S-to-W, which justifies the effectiveness of our method. It can also be observed that most TTA methods (TENT \cite{hu2021fully} and LAME \cite{boudiaf2022parameter}) are outperformed by CTTA methods, which reveals the importance of avoiding catastrophic forgetting.
Additionally, previous MM-TTA (i.e., xMUDA and MMTTA) methods can not consistently outperform previous single-modal methods (e.g., EATA \cite{niu2022efficient}), which suggests the sensitivity of multi-modal collaboration toward continual domain shifts. In fact, by intuitively utilizing model-based stochastic restoration, MMTTA-rs can achieve a noticeable improvement of $3.1\%$ and $6.0\%$ on S-to-S and S-to-W, respectively. Yet MMTTA-rs is still outperformed by our method. To justify the effect of the sequence arrangement, we conduct additional experiments by reversing the sequences of S-to-S and S-to-W for TENT, MMTTA-rs, and our CoMAC. While the performance of all methods decreases, the improvement of CoMAC compared to others can be observed across all sequences.

For S-to-S, we notice that our method achieves the second-best performance at the first sequence 01-Spring and then surpasses all previous methods in the following sequences since we utilize CMQs to gradually adapt to the changing target domain without forgetting. Compared with TTA methods, our method consistently prevails with an increasing gap as they struggle with error accumulation. A similar observation can be found at D-O-2 of S-to-W. Interestingly, xMUDA-pl as well as Pslabel perform competitively on S-to-W and outperform CoMAC in D-P-2 mainly because is less challenging and simply filtering pseudo-labels can effectively mitigate the prediction noise. Nevertheless, this performance gap is trivial as our CoMAC surpasses both xMUDA-pl and Pslabel on all other sequences.

\begin{figure*}[t]
\centering
\includegraphics[width=.8\textwidth]{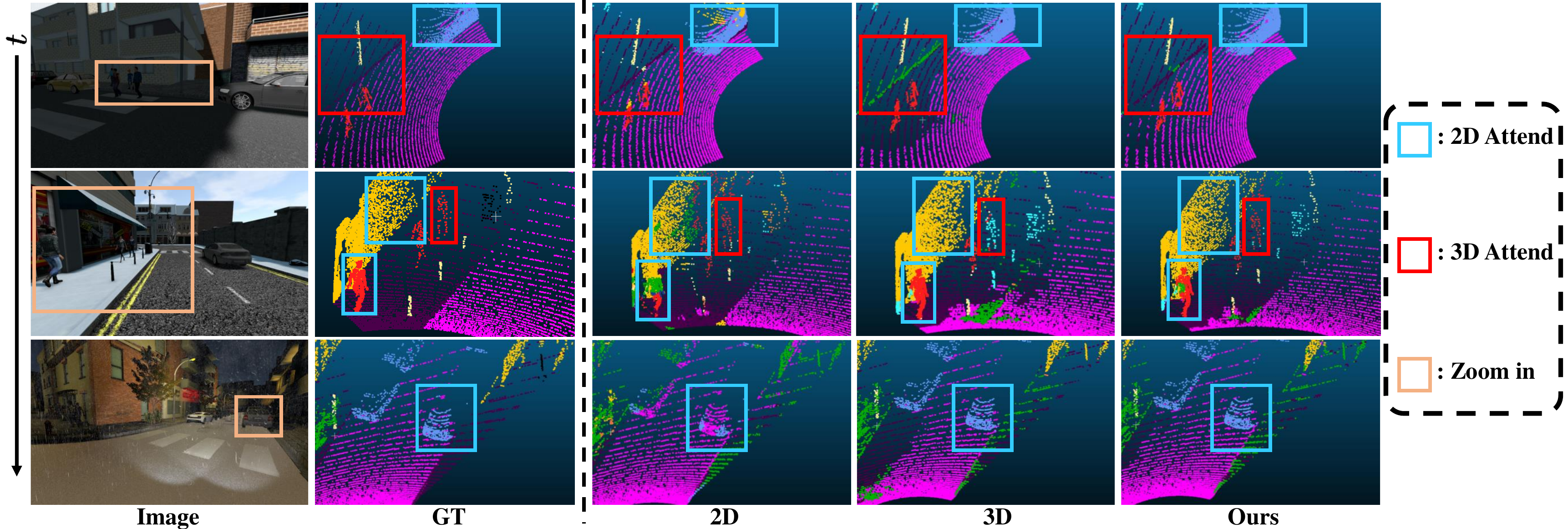}
\caption{Visualization of online segmentation results. Here we present the cross-modal pseudo-labels from xMPF (Ours) as well as the individual prediction from 2D and 3D teachers in comparison with the ground truth (GT).}
\label{Fig:Exp_Visual}
\vspace{-8pt}
\end{figure*}



\subsection{Ablation Studies}\label{Section: Exp_Ablation}
In this section, we provide our ablation studies on S-to-S to justify our design of CoMAC. 

{\setlength{\parindent}{0cm}
\textbf{Utilization of raw and augmented input.} In Sec.~\ref{Section: Method-iMPA}, a key hypothesis is that while the augmentation-average prediction can be a potential reliable alternative, it may introduce inductive bias to the result. To justify this assumption, we conduct experiments by simply using the pure raw or augmentation-average prediction as the intra-modal prediction as in Eq.~\ref{Equations:iMPA-end}. As shown in Fig.~\ref{Fig:Exp_Ablation}(a), both variants achieve inferior performance compared to CoMAC and these performance gaps are consistent across most sequences. Specifically, using pure augmentation-average predictions lead to serious degradation of $5.8\%$, which justifies the necessity of mitigating the inductive noise brought by test-time augmentations and our motivation for iMPA.
}

{\setlength{\parindent}{0cm}
\textbf{Weighting mechanisms of iMPA and xMPF.} The proposed iMPA and xMPF utilize centroid-based weighting mechanisms to attend to the reliable modality. To justify their effectiveness, we compared CoMAC with variants by replacing either Eq.~\ref{Equations:iMPA-end} or Eq.~\ref{Equation:Weight_xMPF} with a simple average fusion (indicated as w/o iMPA and w/o xMPF). As presented in Fig.~\ref{Fig:Exp_Ablation}(b), CoMAC outperforms both variants by more than $2.5\%$. This performance gap justifies that both iMPA and xMPF play an important part in noise suppression.
}

{\setlength{\parindent}{0cm}
\textbf{Updating mechanisms of CMQs.} The goal of CMQs is to achieve a good balance between adaptation and knowledge preservation. To justify the role of CMQs, we compared CoMAC with variants without enqueuing target features or restoring pseudo-source features, respectively. For the variant without enqueuing target features, the corresponding class-wise centroids are identical to the source centroids across the whole testing process. As in Fig.~\ref{Fig:Exp_Ablation}(c), disabling enqueuing target features (w/o Update) causes a significant performance decrease of $6.1\%$, which justifies the necessity of adapting class-wise centroids for the validity of iMPA and xMPF. On the other hand, while disabling restoring pseudo-source features (w/o Restore) leads to a quicker adaptation at the first two sequences, it eventually causes catastrophic forgetting after the third sequence, resulting in a noticeable gap of $8.4\%$. Overall, the performance gain brought by CMQs justifies their effectiveness.
}

\begin{table}[t]
\centering
\resizebox{.87\linewidth}{!}{
	\begin{tabular}{m{4em}<{\centering} m{0.25em} | m{2.6em}<{\centering} m{2.6em}<{\centering} m{2.6em}<{\centering} m{2.6em}<{\centering} m{2.6em}<{\centering} | m{2.6em}<{\centering}}
		\hline
		\multicolumn{2}{c|}{\multirow{2}{*}{CoMAC}} & \multicolumn{6}{c}{No. of 2D Aug} \\
		\cline{3-8}
		& & 1 & 2 & 3 & 4 & 5 & Avg \\
		\hline
		\multirow{6}{*}{\shortstack{No. of \\ 3D Aug}} & \multicolumn{1}{|c|}{1} & 33.0 & 33.1 & 33.2 & 33.2 & 33.3 & 33.2 \\
		& \multicolumn{1}{|c|}{2} & 34.1 & 34.2 & 34.3 & 34.4 & 34.4 & 34.3\\
		& \multicolumn{1}{|c|}{3} & 35.2 & 35.2 & 35.4 & 35.3 & 35.5 & 35.3\\
		& \multicolumn{1}{|c|}{4} & 36.4 & 36.4 & 36.5 & 36.5 & 36.5 & 36.5\\
		& \multicolumn{1}{|c|}{5} & 36.4 & 36.5 & \textbf{36.7} & 36.6 & \textbf{36.7} & 36.6 \\
		\cline{2-8}
		& \multicolumn{1}{|c|}{Avg} & 35.0 & 35.1 & 35.2 & 35.2 & 35.3 & 35.2\\
		\hline
\end{tabular}}
\smallskip
\caption{Augmentation analysis of CoMAC. CoMAC is evaluated with different numbers of augmentations.}
\label{Table: Exp_aug}
\vspace{-5pt}
\end{table}

\begin{table}[t]
\centering
\resizebox{.97\linewidth}{!}{
	\begin{tabular}{c | c c | c | c c | c | c c | c }
		\hline
		\multirow{5}{*}{\rotatebox{90}{CoMAC}} & $p_{rs}$ & $(\tau_{cf},\lambda_{cts})$ & Avg & $\tau_{cf}$ & $(p_{rs},\lambda_{cts})$ & Avg & $\lambda_{cts}$ & $(p_{rs},\tau_{cf})$ &  Avg  \\
		\cline{2-10}
		& 0.5 & (0.8, 1.0) & 36.6 & 0.8 & (0.5, 1.0) & 36.6 & 1.0 & (0.5, 0.8)  & 36.6 \\
		\cline{2-10}
		& 0.3 & - & 34.5 & 0.0 & - & 34.8 & 0.0 & - & 35.8\\
		& 0.7 & - & 36.4 & 0.5 & - & 36.5 & 0.01 & - & 36.4\\
		& 1.0 & - & 36.3 & 0.9 & - & 36.4 & 0.1 & - & 36.4\\
		\hline
\end{tabular}}
\smallskip
\caption{Sensitivity analysis of CoMAC. Here ``xM'' is the softmax-average of 2D and 3D predictions. The parameter value with ``-'' indicates the same value as default settings.}
\label{Table: Exp_sensity}
\vspace{-5pt}
\end{table}

{\setlength{\parindent}{0cm}
\textbf{Number of Augmentations.} To investigate the effect of augmentations, we conduct a grid search by utilizing different numbers of 2D and 3D augmentations. Specifically, we adopt resize and rotation as 2D and 3D augmentation, respectively. The factor range for 2D scaling is $[0.5,1)$ while 3D one is set to $(0^{\circ}, 360^{\circ})$, where the list of augmentations is computed as evenly spaced factors given the number of augmentations. As shown in Table~\ref{Table: Exp_aug}, increasing the number of either 2D or 3D augmentations leads to performance improvement, while the 2D improvement is less noticeable compared to 3D probably due to the inductive bias brought by 2D augmentations. While all settings perform competitively, we adopt the combination of three 2D augmentations and five 3D augmentations to achieve the best result.
}

{\setlength{\parindent}{0cm}
\textbf{Sensitivity analysis.} We perform sensitivity analysis over the hyper-parameters of CoMAC as shown in Table~\ref{Table: Exp_sensity}. There are some extreme cases that can lead to the performance degeneration of CoMAC. For restoring rate $p_{rs}$, the performance relatively drops about $5.7\%$ when $p_{rs}=0.3$, indicating the restoring rate is too low to prevent forgetting. Nevertheless, when $p_{rs} \geq 0.5$, CoMAC is robust to the arbitrary choice of $p_{rs}$. Similar observations can be found when $\tau_{cf}=0.0$ or $\lambda_{cts}=0.0$, where the performance decreases relatively by $4.9\%$ and $2.2\%$. This performance degeneration indicates the importance of confident feature filtering and class-wise contrastive loss. Overall, when $\tau_{cf} \geq 0.0$ and $\lambda_{cts} \geq 0.01$, the performance of CoMAC is also robust to the hyper-parameter settings, falling into a relative margin of $0.5\%$. Note that all settings of CoMAC surpass the previous SOTA method EATA \cite{niu2022efficient} by more than $4.8\%$.
}

{\setlength{\parindent}{0cm}
\textbf{Visualization of segmentation.} To further justify the effectiveness of our proposed CoMAC, we provide some visualization results as shown in Fig.~\ref{Fig:Exp_Visual}. Specifically, we visualize the 2D and 3D predictions from teacher models and our cross-modal pseudo-labels from xMPF. Compared to single-modal predictions, our cross-modal pseudo-labels can achieve better segmentation results by attending to the reliable modality, such as the first row in Fig.~\ref{Fig:Exp_Visual}, where the sidewalk prediction attends more to the 2D image while the car prediction attends to the reliable 3D point cloud.
}

\section{Conclusion}
In this paper, we present a new task, named multi-modal continual test-time adaptation (MM-CTTA) for 3D semantic segmentation. We further propose a novel method called CoMAC that tackles MM-CTTA from two perspectives. On one hand, reliable cross-modal pseudo-labels are generated by iMPA and xMPF by adaptively attending to the more reliable modality in a dual-stage manner. On the other hand, CMQs are proposed to leverage pseudo-source features and reliable target features to perform adaptation without catastrophic forgetting. We introduce two new benchmarks for MM-CTTA and our methods outperform previous works by a noticeable margin in both benchmarks. In the future, we hope both our method and benchmarks can facilitate the exploration of MM-CTTA and promote the development of reliable multi-modal systems in altering environments.

\section*{Appendix}

\textit{In this appendix, we provide more details about our proposed CoMAC. \textbf{Firstly}, we present more details about our proposed benchmarks, including the generation of synthetic point clouds of Synthia \cite{ros2016synthia} and class mapping for both benchmarks. \textbf{Secondly}, we introduce the implementation details of the baselines we compared with CoMAC. \textbf{Thirdly}, the implementation details of CoMAC are included and we provide additional ablation studies to justify our implementation. \textbf{Fourthly}, we present more visualization comparisons between CoMAC and previous SOTA methods to show that our CoMAC can outperform prior works during the continual adaptation. \textbf{Last but not least}, we thoroughly compare our CoMAC with previous works to justify the novelty of our work.}

\subsection*{Benchmark Details}
In Sec.~4.1, we propose two new benchmarks for the exploration of MM-CTTA, including SemanticKITTI-to-Synthia (S-to-S) and SemanticKITTI-to-Waymo (S-to-W). Unlike conventional UDA benchmarks for image semantic segmentation which usually treat simulation datasets (in this case, Synthia \cite{ros2016synthia}) as the source domain, we regard Synthia as the target domain for S-to-S since it contains more various environmental conditions compared to most real-life datasets \cite{behley2019semantickitti,caesar2020nuscenes,shin2022mm}, which enables us to construct a more challenging MM-CTTA benchmark. Although there exist some real-life datasets recorded under adverse conditions \cite{Yu_2020_CVPR,sakaridis2021acdc}, they usually contain single-modal information and therefore are not applicable under our settings. 

\begin{figure}
\centering
\includegraphics[width=\linewidth]{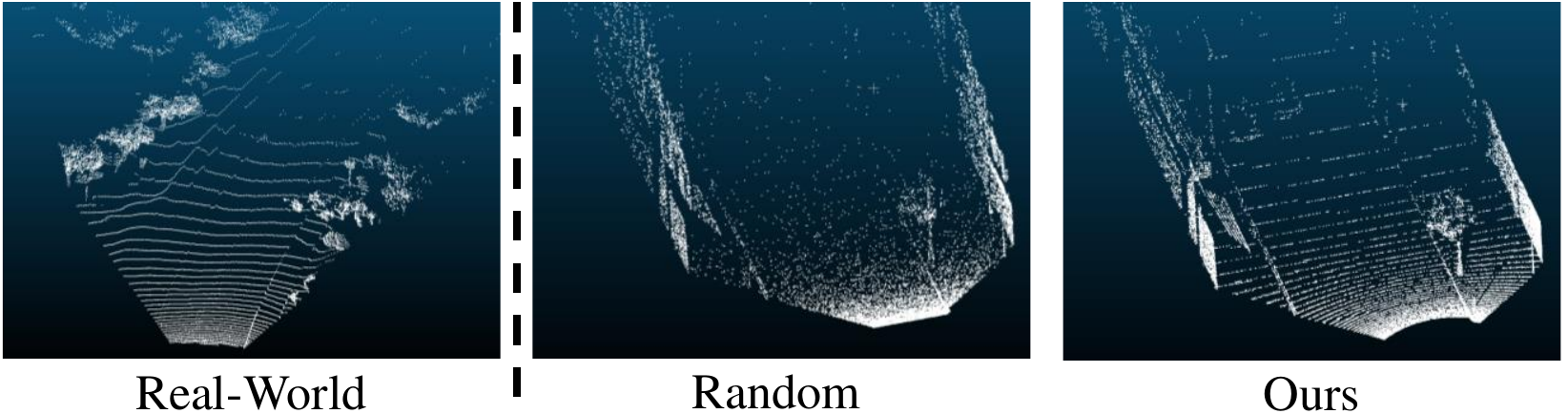}
\caption{Comparison of the real-world point cloud, randomly sampled depth pixels as point clouds in MMTTA \cite{shin2022mm} and ours.}
\label{Fig:point}
\end{figure}

{\noindent
\textbf{Point cloud generation for Synthia}. For S-to-S, we generate synthetic 3D point clouds from depth images as mentioned in Sec.~4.1. More specifically, the 3D point clouds are extracted by mimicking the point clouds from a 64-line LiDAR with a vertical field-of-view (FOV) of $30^\circ$. Given the 2D depth images, we first project the 2D depth pixels back to the 3D world based on the official intrinsic matrix. Subsequently, we limit the vertical FOV of pixels by the range $[-15^{\circ},15^{\circ}]$ and simulate the characteristic of LiDAR channels by vertically selecting 64 lines of pixels that evenly divide the FOV. After filtering out the points located too far ($>100$m), the simulated 3D points are then randomly sampled from each line, leading to a total of 20k points for each frame. As shown in Fig.~\ref{Fig:point}, compared to the previous work MMTTA \cite{shin2022mm} that randomly samples pixels without considering the structural characteristic of points, our simulated point clouds are more similar to those collected by real-world LiDARs.
}

{\noindent
\textbf{Class mapping}. To address the label misalignment between the source-domain and target-domain datasets, we utilize a similar label mapping strategy as \cite{jaritz2020xmuda} to form an 11-class mapping for each benchmark. Most of the classes are identical to \cite{jaritz2020xmuda} as shown in Table.~\ref{Table:Synthia_mapping} and Table.~\ref{Table:Waymo_mapping}. For SemanticKITTI-to-Synthia, we ignore the class ``sky'' of Synthia since it is neither included in SemanticKITTI nor a common class in 3D segmentation. For SemanticKITTI-to-Waymo, we introduce the new class ``trunk'' which is not included in \cite{jaritz2020xmuda} considering its important role in other downstream tasks (e.g., localization \cite{chen2019suma++} and map construction \cite{mccormac2017semanticfusion,wang2021multi}). Both mappings form a segmentation task of 11 classes.
}

\begin{table*}[t]
\centering
\resizebox{.75\linewidth}{!}{
\begin{tabular}{m{5.5em} | m{18em} | m{18em}}
\hline
\hline
S-to-S Class & SemanticKITTI classes & Synthia classes\\
\hline
car & car, moving-car, truck, moving-truck & car\\
bike & bicycle, motorcycle, bicyclist, motorcyclist, moving-bicyclist, moving-motorcyclist & bicycle \\
person & person, moving-person & pedestrian \\
road & road, lane-marking, parking &  road, lanemarking \\
sidewalk & sidewalk & sidewalk \\
building & building & building \\
nature & vegetation, trunk, terrain & vegetation \\
poles & pole & pole \\
fence & fence & fence \\
traffic-sign & traffic-sign & traffic-sign \\
other-objects & other-object & traffic-light \\
\hline
\hline
\end{tabular}
}
\caption{Class mapping of the benchmark SemanticKITTI-to-Synthia.}
\label{Table:Synthia_mapping}
\end{table*}

\begin{table*}[t]
\centering
\resizebox{.75\linewidth}{!}{
\begin{tabular}{m{5.5em} | m{18em} | m{18em}}
\hline
\hline
S-to-W Class & SemanticKITTI classes & Waymo classes\\
\hline
car & car, moving-car, truck, moving-truck & car, truck, other-vehicle, bus\\
bike & bicycle, motorcycle, bicyclist, motorcyclist, moving-bicyclist, moving-motorcyclist & bicycle, motorcycle, bicyclist, motorcyclist \\
person & person, moving-person & pedestrian \\
road & road, lane-marking, parking &  road, lane-marker \\
sidewalk & sidewalk & sidewalk, curb, walkable \\
building & building & building \\
nature & vegetation, terrain & vegetation \\
poles & pole & pole \\
trunk & trunk & tree-trunk \\
traffic-sign & traffic-sign & sign \\
other-objects & other-object & other-ground, construction-cone, traffic-light \\
\hline
\hline
\end{tabular}
}
\caption{Class mapping of the benchmark SemanticKITTI-to-Waymo.}
\label{Table:Waymo_mapping}
\end{table*}

\subsection*{Details of pre-trained models and baselines}
Here we present implementation details of the pre-training stage and baseline methods in our work. For all methods, we conduct a search for learning rates and report the best results.

{\noindent
\textbf{Pre-training on SemanticKITTI.} Similar to previous works \cite{jaritz2020xmuda,shin2022mm}, we follow the official code of \cite{jaritz2020xmuda} to pre-train both 2D and 3D networks on SemanticKITTI \cite{behley2019semantickitti}. Specifically, the optimizers for both networks are Adam \cite{kingma2014adam} with $\beta=0.9$ and $\beta=0.999$, while the base learning rate is set to 0.001 with a batch size of 8 for both networks. The voxel scale and full scale of SCN are set to 20 and 4096, respectively. The total pre-training epoch is set to 100,000, where the learning rate shrinks ten-fold at the epochs 8,000 and 9,000. For 2D augmentations, we utilize a bottom crop of the size $(480, 302)$ followed by a random horizontal flip with a probability of 0.5 as well as color jittering of $(0.4, 0.4, 0.4)$. On the other hand, 3D augmentations during the pre-training stage include a noisy rotation of 0.1, a random flip of the y-axis with a probability of 0.5, a random z-rotation within the range $[0, 360^\circ]$, and a random translation within the receptive field of SCN. The pre-trained models are then utilized as the initial state for all methods.
}

{\noindent
\textbf{TTA methods.} For TTA methods, we compared our CoMAC with self-training with pseudo-labels (Pslabel), TENT \cite{wang2020tent}, and LAME \cite{boudiaf2022parameter}. Specifically, hard thresholding is introduced in Pslabel to mitigate the side-effect of noisy pseudo-labels, where pseudo-labels with confidence scores less than the median score of their predicted class are filtered out. For TENT \cite{wang2020tent}, the implementation follows the official code. For LAME \cite{boudiaf2022parameter}, we implement the variant with the linear affinity matrix instead since either kNN or rbf affinity is too expensive to compute for segmentation. 
}

{\noindent
\textbf{MM-TTA methods.} xMUDA \cite{jaritz2020xmuda} with pseudo-labels (xMUDA-pl) and MMTTA \cite{shin2022mm} are selected as representatives of MM-TTA methods. Specifically, xMUDA-pl utilizes the same pseudo-label filtering strategy as Pslabel, and we utilize the cross-modal consistency learning of xMUDA in a test-time manner. For MMTTA, we self-implement the proposed method since it does not come with any official code. As we mentioned in Sec. 4.2, we additionally introduce an MM-CTTA variant of MMTTA simply by integrating MMTTA with the model-based stochastical restoration. Specifically, the restoring rate is set to 0.01 for MMTTA-rs for both benchmarks.
}

{\noindent
\textbf{CTTA methods.} In this work, CoTTA \cite{wang2022continual} and EATA \cite{niu2022efficient} are the reason works that focus on the CTTA problem. Both CoTTA and EATA follow the official implementations. For CoTTA which is also based on test-time augmentations, we apply the same z-rotation as 3D augmentations for the adaptation of the 3D backbone. For EATA, we adopt the first sequence of each benchmark to train the Fisher regularizer. 
}

\begin{figure*}[t]
\centering
\subfloat[][]{\includegraphics[width=0.32\textwidth]{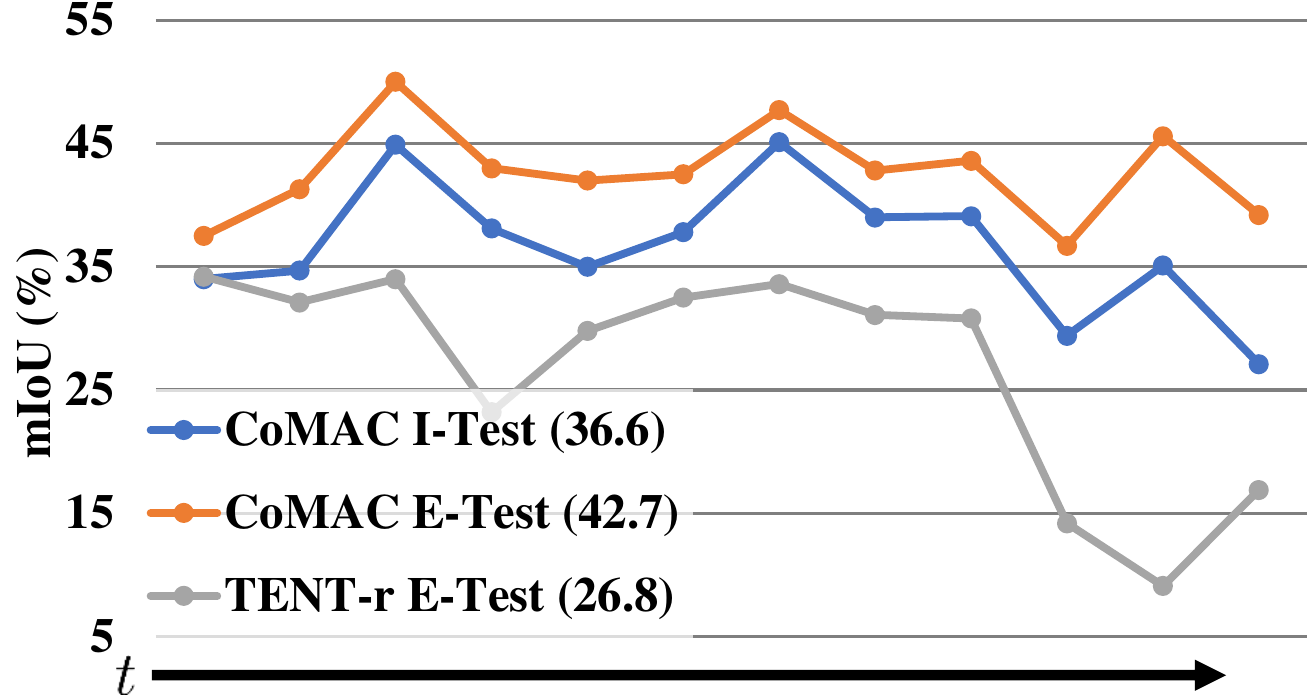}\label{Fig:mp_vs_op}}
\hspace{0.1em}
\subfloat[][]{\includegraphics[width=0.32\textwidth]{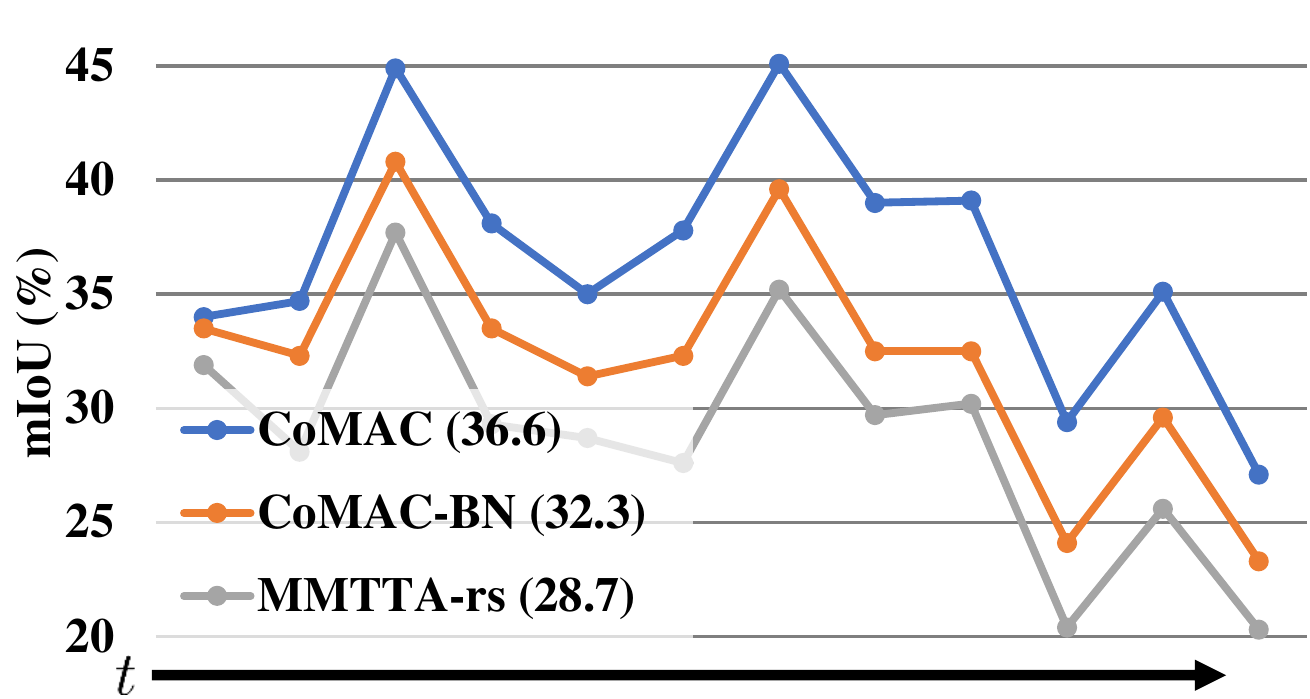}\label{Fig:bn}}
\hspace{0.1em}
\subfloat[][]{\includegraphics[width=0.32\textwidth]{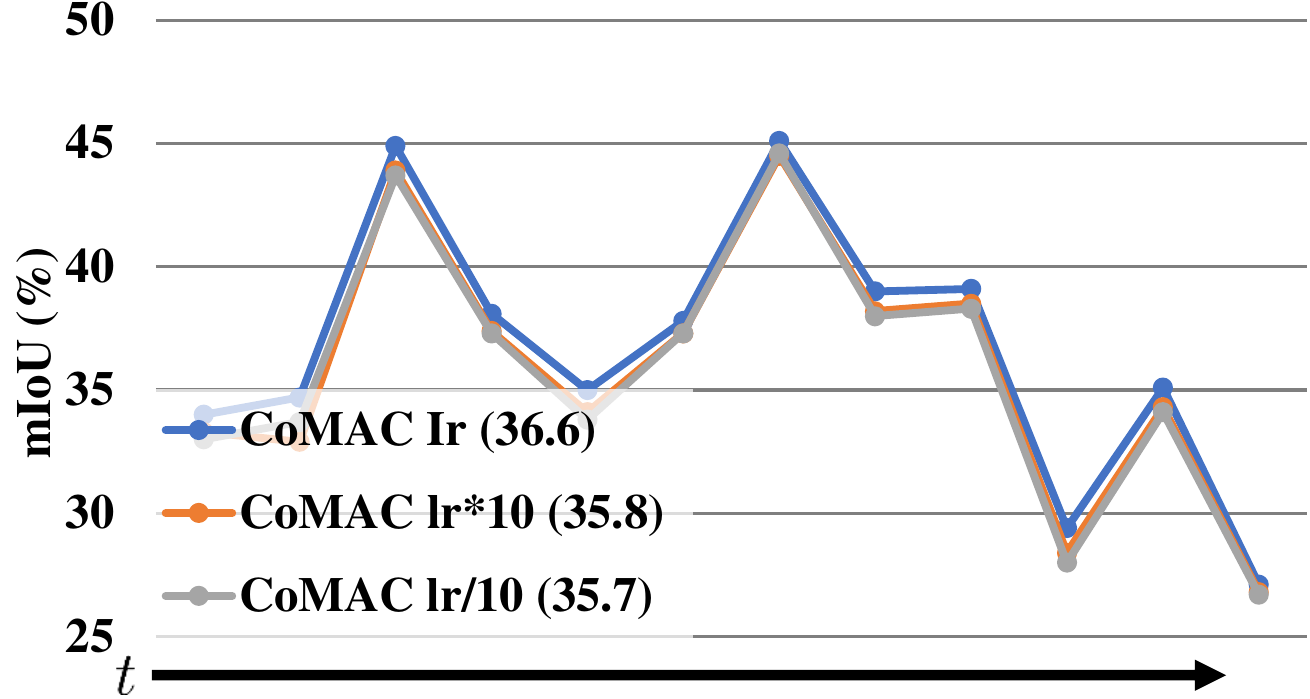}\label{Fig:lr}}
\caption{Additional ablation studies to justify the effectiveness of CoMAC. Specifically, Fig.~\ref{Fig:mp_vs_op} illustrates the performance comparison between testing immediately (I-Test) and testing at the end of each sequence (E-Test), while Fig.~\ref{Fig:bn} compares CoMAC with its variant which only update the parameters of batch normalization layers. Fig.~\ref{Fig:lr} presents the robustness of CoMAC toward different learning rates.}
\label{Fig:Supp_abation}
\vspace{-5pt}
\end{figure*}

\begin{figure*}[t]
\centering
\includegraphics[width=\textwidth]{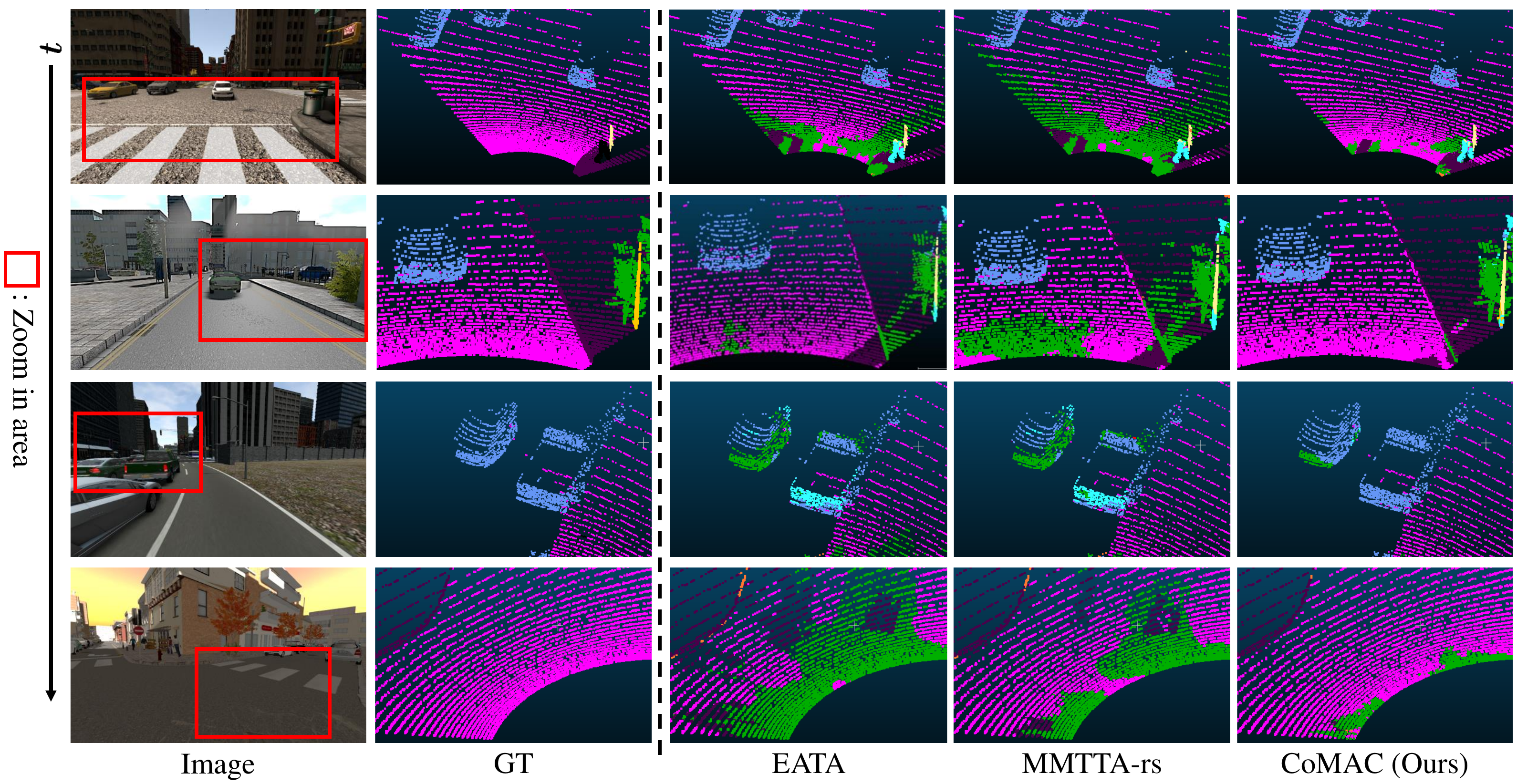}
\caption{Visualization comparison between EATA \cite{niu2022efficient}, MMTTA-rs \cite{shin2022mm}, and CoMAC (ours). Sequences from the top to the bottom are from 02-Summer, 04-Fall, 05-Winter, and 04-Sunset, respectively.}
\label{Fig:visual}
\vspace{-5pt}
\end{figure*}

\subsection*{Implementation Details of CoMAC}
As mentioned in Sec.~3.4, the source class-wise centroids are collected as source knowledge. The source centroids are computed as the average of normalized features of all source training samples based on their labels. During the test-time adaptation, the batch normalization layers utilize the batch statistics of the current iteration for predictions. The optimizers of CoMAC for both networks are identical to the pre-training stage, while the learning rate is set to 1.25E-05. The batch size is set to 1 and the performance is based on the immediate testing results (I-Test) after the encounter of the data, strictly following the one-pass protocol as in previous works \cite{wang2022continual,shin2022mm}. Here CoMAC updates all parameters of both networks. All our experiments are conducted on an NVIDIA RTX 3090. Our code is implemented based on PyTorch \cite{paszke2019pytorch} of version 1.8.1 with CUDA 11.4.

\begin{table*}[t]
\centering
\resizebox{.9\linewidth}{!}{
\begin{tabular}{m{6.5em} | m{4.5em} | m{14em} | m{18em}}
	\hline
	\hline
	Method & Publication & Task & Designs\\
	\hline
	TENT \cite{wang2020tent} & ICLR-21 & 
	Test-Time Adaptation (TTA): inaccessible source data, a \textbf{stationary} target domain with the input of \textbf{a single modality}, adaptation conducted during the testing process. & 
	(i) TENT proposes to update the parameters of batch normalization layers by minimizing the prediction entropy in the target domain; (ii) TENT highlights \textbf{the first TTA protocol} for future exploration.\Tstrut\Bstrut\\
	\hline
	MMTTA \cite{shin2022mm} & CVPR-22 & Multi-Modal Test-Time Adaptation (MM-TTA): inaccessible source data,  a \textbf{stationary} target domain with the input of \textbf{multiple modalities}, adaptation conducted during the testing process. & 
	(i) MMTTA generates reliable intra-modal pseudo labels by combining predictions from student and teacher models. (ii) The inter-modal pseudo labels are generated as the weighted sum of intra-modal predictions based on \textbf{student-teacher consistency}.\Tstrut\Bstrut\\
	\hline
	CoTTA \cite{wang2022continual} & CVPR-22 & Continual Test-Time Adaptation (CTTA): inaccessible source data,  a \textbf{continually changing} target domain with the input of \textbf{a single modality}, adaptation conducted during the testing process. &
	(i) CoTTA estimates the reliability of teacher predictions through confidence score, where \textbf{hard-selected} unreliable predictions are substituted by augmentation-average for noise suppression. (ii) Parameters of networks are stochastically restored to avoid catastrophic forgetting.\Tstrut\Bstrut\\
	\hline
	CoMAC (Ours) & - & Multi-Modal Continual Test-Time Adaptation (MM-CTTA): inaccessible source data,  a \textbf{continually changing} target domain with the input of \textbf{multiple modalities}, adaptation conducted during the testing process. &
	(i) CoMAC utilizes iMPA as \textbf{a centroid-based adaptive fusion} between augmentation-average and raw predictions to mitigate the potential inductive bias. (ii) CoMAC \textbf{attends to reliable modality} through xMPF for cross-modal noise suppression. (iii) The centroids are actively updated for adaptation while \textbf{stochastically revisiting pseudo-source features} as source knowledge to avoid forgetting.\Tstrut\Bstrut\\
	\hline
	\hline
\end{tabular}
}
\caption{Comparison between our CoMAC and previous methods.}
\label{Table:Method_Comp}
\end{table*}

To justify the effectiveness of CoMAC under different evaluation protocols and settings, we additionally conduct experiments where CoMAC is evaluated at the end of each sequence. Specifically, as shown in Fig.~\ref{Fig:mp_vs_op}, testing at the end of each sequence (E-Test) results in a noticeable improvement of $6.1\%$ compared to the I-Test version of CoMAC. Here we also present an E-Test version of TENT with auto-reset at the beginning of each sequence (in short, TENT-r E-Test) as a fair comparison, while both I-Test and E-Test versions of CoMAC surpass TENT-r E-Test by more than $9.8\%$, which justifies the effectiveness of CoMAC. Fig.~\ref{Fig:bn} presents the performance of the CoMAC variant which only updates the parameters of batch normalization layers, outperforming MMTTA-rs which also only updates batch normalization layers by about $3.6\%$. As for the learning rate, Fig.~\ref{Fig:lr} justifies the robustness of CoMAC toward learning rates, falling into a relative margin of $2.4\%$.

\subsection*{Visualization Comparison}
To intuitively justify the effectiveness of our CoMAC, we present more visualization results of our CoMAC in comparison with previous SOTA methods, including EATA \cite{niu2022efficient} and MMTTA-rs \cite{shin2022mm}. Specifically, we present the visualization comparison from different sequences as illustrated in Fig.~\ref{Fig:visual}, where our CoMAC generates more accurate predictions compared to both EATA and MMTTA-rs.

\subsection*{Comparison with Previous (C)TTA and MM-TTA Methods}
In this work, we propose CoMAC to tackle MM-CTTA for 3D semantic segmentation as an extension of CTTA. To highlight our novelty, we compare our proposed CoMAC with previous TTA (TENT \cite{wang2020tent}), CTTA (CoTTA \cite{wang2022continual}), and MM-TTA (MMTTA \cite{shin2022mm}) methods. Specifically, we list out their specific tasks and designs as shown in Table.~\ref{Table:Method_Comp}.

{\small
\bibliographystyle{ieee_fullname}
\bibliography{egbib}

\begin{thebibliography}{10}\itemsep=-1pt

\bibitem{bayoudh2021survey}
Khaled Bayoudh, Raja Knani, Fay{\c{c}}al Hamdaoui, and Abdellatif Mtibaa.
\newblock A survey on deep multimodal learning for computer vision: advances,
  trends, applications, and datasets.
\newblock {\em The Visual Computer}, pages 1--32, 2021.

\bibitem{behley2019semantickitti}
Jens Behley, Martin Garbade, Andres Milioto, Jan Quenzel, Sven Behnke, Cyrill
  Stachniss, and Jurgen Gall.
\newblock Semantickitti: A dataset for semantic scene understanding of lidar
  sequences.
\newblock In {\em Proceedings of the IEEE/CVF International Conference on
  Computer Vision}, pages 9297--9307, 2019.

\bibitem{boudiaf2022parameter}
Malik Boudiaf, Romain Mueller, Ismail Ben~Ayed, and Luca Bertinetto.
\newblock Parameter-free online test-time adaptation.
\newblock In {\em Proceedings of the IEEE/CVF Conference on Computer Vision and
  Pattern Recognition}, pages 8344--8353, 2022.

\bibitem{bultmann2023real}
Simon Bultmann, Jan Quenzel, and Sven Behnke.
\newblock Real-time multi-modal semantic fusion on unmanned aerial vehicles
  with label propagation for cross-domain adaptation.
\newblock {\em Robotics and Autonomous Systems}, 159:104286, 2023.

\bibitem{caesar2020nuscenes}
Holger Caesar, Varun Bankiti, Alex~H Lang, Sourabh Vora, Venice~Erin Liong,
  Qiang Xu, Anush Krishnan, Yu Pan, Giancarlo Baldan, and Oscar Beijbom.
\newblock nuscenes: A multimodal dataset for autonomous driving.
\newblock In {\em Proceedings of the IEEE/CVF Conference on Computer Vision and
  Pattern Recognition}, pages 11621--11631, 2020.

\bibitem{chen2019progressive}
Chaoqi Chen, Weiping Xie, Wenbing Huang, Yu Rong, Xinghao Ding, Yue Huang,
  Tingyang Xu, and Junzhou Huang.
\newblock Progressive feature alignment for unsupervised domain adaptation.
\newblock In {\em Proceedings of the IEEE/CVF Conference on Computer Vision and
  Pattern Recognition}, pages 627--636, 2019.

\bibitem{chen2022contrastive}
Dian Chen, Dequan Wang, Trevor Darrell, and Sayna Ebrahimi.
\newblock Contrastive test-time adaptation.
\newblock In {\em Proceedings of the IEEE/CVF Conference on Computer Vision and
  Pattern Recognition}, pages 295--305, 2022.

\bibitem{chen2019suma++}
Xieyuanli Chen, Andres Milioto, Emanuele Palazzolo, Philippe Giguere, Jens
  Behley, and Cyrill Stachniss.
\newblock Suma++: Efficient lidar-based semantic slam.
\newblock In {\em 2019 IEEE/RSJ International Conference on Intelligent Robots
  and Systems (IROS)}, pages 4530--4537. IEEE, 2019.

\bibitem{EasMasWilSch21}
C. Eastwood, I. Mason, C. Williams, and B. Sch{\"o}lkopf.
\newblock Source-free adaptation to measurement shift via bottom-up feature
  restoration.
\newblock In {\em 10th International Conference on Learning Representations
  (ICLR)}, Apr. 2022.

\bibitem{fleuret2021uncertainty}
Francois Fleuret et~al.
\newblock Uncertainty reduction for model adaptation in semantic segmentation.
\newblock In {\em Proceedings of the IEEE/CVF Conference on Computer Vision and
  Pattern Recognition}, pages 9613--9623, 2021.

\bibitem{gan2022decorate}
Yulu Gan, Xianzheng Ma, Yihang Lou, Yan Bai, Renrui Zhang, Nian Shi, and Lin
  Luo.
\newblock Decorate the newcomers: visual domain prompt for continual test time
  adaptation.
\newblock {\em arXiv preprint arXiv:2212.04145}, 2022.

\bibitem{gongnote}
Taesik Gong, Jongheon Jeong, Taewon Kim, Yewon Kim, Jinwoo Shin, and Sung-Ju
  Lee.
\newblock Note: Robust continual test-time adaptation against temporal
  correlation.
\newblock In {\em Advances in Neural Information Processing Systems}, 2022.

\bibitem{goyaltest}
Sachin Goyal, Mingjie Sun, Aditi Raghunathan, and J~Zico Kolter.
\newblock Test time adaptation via conjugate pseudo-labels.
\newblock In {\em Advances in Neural Information Processing Systems}, 2022.

\bibitem{graham20183d}
Benjamin Graham, Martin Engelcke, and Laurens Van Der~Maaten.
\newblock 3d semantic segmentation with submanifold sparse convolutional
  networks.
\newblock In {\em Proceedings of the IEEE Conference on Computer Vision and
  Pattern Recognition}, pages 9224--9232, 2018.

\bibitem{he2020momentum}
Kaiming He, Haoqi Fan, Yuxin Wu, Saining Xie, and Ross Girshick.
\newblock Momentum contrast for unsupervised visual representation learning.
\newblock In {\em Proceedings of the IEEE/CVF Conference on Computer Vision and
  Pattern Recognition}, pages 9729--9738, 2020.

\bibitem{he2016deep}
Kaiming He, Xiangyu Zhang, Shaoqing Ren, and Jian Sun.
\newblock Deep residual learning for image recognition.
\newblock In {\em Proceedings of the IEEE Conference on Computer Vision and
  Pattern Recognition}, pages 770--778, 2016.

\bibitem{hoffman2018cycada}
Judy Hoffman, Eric Tzeng, Taesung Park, Jun-Yan Zhu, Phillip Isola, Kate
  Saenko, Alexei Efros, and Trevor Darrell.
\newblock Cycada: Cycle-consistent adversarial domain adaptation.
\newblock In {\em International Conference on Machine Learning}, pages
  1989--1998. Pmlr, 2018.

\bibitem{hu2021fully}
Minhao Hu, Tao Song, Yujun Gu, Xiangde Luo, Jieneng Chen, Yinan Chen, Ya Zhang,
  and Shaoting Zhang.
\newblock Fully test-time adaptation for image segmentation.
\newblock In {\em Medical Image Computing and Computer Assisted
  Intervention--MICCAI 2021: 24th International Conference, Strasbourg, France,
  September 27--October 1, 2021, Proceedings, Part III 24}, pages 251--260.
  Springer, 2021.

\bibitem{jaritz2020xmuda}
Maximilian Jaritz, Tuan-Hung Vu, Raoul~de Charette, Emilie Wirbel, and Patrick
  Perez.
\newblock xmuda: Cross-modal unsupervised domain adaptation for 3d semantic
  segmentation.
\newblock In {\em Proceedings of the IEEE/CVF Conference on Computer Vision and
  Pattern Recognition}, June 2020.

\bibitem{khosla2020supervised}
Prannay Khosla, Piotr Teterwak, Chen Wang, Aaron Sarna, Yonglong Tian, Phillip
  Isola, Aaron Maschinot, Ce Liu, and Dilip Krishnan.
\newblock Supervised contrastive learning.
\newblock {\em Advances in Neural Information Processing Systems},
  33:18661--18673, 2020.

\bibitem{kingma2014adam}
Diederik~P Kingma and Jimmy Ba.
\newblock Adam: A method for stochastic optimization.
\newblock {\em arXiv preprint arXiv:1412.6980}, 2014.

\bibitem{kundu2021generalize}
Jogendra~Nath Kundu, Akshay Kulkarni, Amit Singh, Varun Jampani, and
  R~Venkatesh Babu.
\newblock Generalize then adapt: Source-free domain adaptive semantic
  segmentation.
\newblock In {\em Proceedings of the IEEE/CVF International Conference on
  Computer Vision}, pages 7046--7056, 2021.

\bibitem{kundu2022balancing}
Jogendra~Nath Kundu, Akshay~R Kulkarni, Suvaansh Bhambri, Deepesh Mehta,
  Shreyas~Anand Kulkarni, Varun Jampani, and Venkatesh~Babu Radhakrishnan.
\newblock Balancing discriminability and transferability for source-free domain
  adaptation.
\newblock In {\em International Conference on Machine Learning}, pages
  11710--11728. PMLR, 2022.

\bibitem{liang2020we}
Jian Liang, Dapeng Hu, and Jiashi Feng.
\newblock Do we really need to access the source data? source hypothesis
  transfer for unsupervised domain adaptation.
\newblock In {\em International Conference on Machine Learning}, pages
  6028--6039. PMLR, 2020.

\bibitem{lim2023ttn}
Hyesu Lim, Byeonggeun Kim, Jaegul Choo, and Sungha Choi.
\newblock Ttn: A domain-shift aware batch normalization in test-time
  adaptation.
\newblock {\em arXiv preprint arXiv:2302.05155}, 2023.

\bibitem{liu2021adversarial}
Wei Liu, Zhiming Luo, Yuanzheng Cai, Ying Yu, Yang Ke, Jos{\'e}~Marcato Junior,
  Wesley~Nunes Gon{\c{c}}alves, and Jonathan Li.
\newblock Adversarial unsupervised domain adaptation for 3d semantic
  segmentation with multi-modal learning.
\newblock {\em ISPRS Journal of Photogrammetry and Remote Sensing},
  176:211--221, 2021.

\bibitem{liu2021ttt++}
Yuejiang Liu, Parth Kothari, Bastien Van~Delft, Baptiste Bellot-Gurlet, Taylor
  Mordan, and Alexandre Alahi.
\newblock Ttt++: When does self-supervised test-time training fail or thrive?
\newblock {\em Advances in Neural Information Processing Systems},
  34:21808--21820, 2021.

\bibitem{liu2021source}
Yuang Liu, Wei Zhang, and Jun Wang.
\newblock Source-free domain adaptation for semantic segmentation.
\newblock In {\em Proceedings of the IEEE/CVF Conference on Computer Vision and
  Pattern Recognition}, pages 1215--1224, 2021.

\bibitem{mccloskey1989catastrophic}
Michael McCloskey and Neal~J Cohen.
\newblock Catastrophic interference in connectionist networks: The sequential
  learning problem.
\newblock In {\em Psychology of Learning and Motivation}, volume~24, pages
  109--165. Elsevier, 1989.

\bibitem{mccormac2018fusion}
John Mccormac, Ronald Clark, Michael Bloesch, Andrew Davison, and Stefan
  Leutenegger.
\newblock Fusion++: volumetric object-level slam.
\newblock In {\em 2018 International Conference on 3D Vision (3DV)}, pages
  32--41, 2018.

\bibitem{mccormac2017semanticfusion}
John McCormac, Ankur Handa, Andrew Davison, and Stefan Leutenegger.
\newblock Semanticfusion: Dense 3d semantic mapping with convolutional neural
  networks.
\newblock In {\em 2017 IEEE International Conference on Robotics and Automation
  (ICRA)}, pages 4628--4635. IEEE, 2017.

\bibitem{niu2022efficient}
Shuaicheng Niu, Jiaxiang Wu, Yifan Zhang, Yaofo Chen, Shijian Zheng, Peilin
  Zhao, and Mingkui Tan.
\newblock Efficient test-time model adaptation without forgetting.
\newblock In {\em International Conference on Machine Learning}, pages
  16888--16905. PMLR, 2022.

\bibitem{paszke2019pytorch}
Adam Paszke, Sam Gross, Francisco Massa, Adam Lerer, James Bradbury, Gregory
  Chanan, Trevor Killeen, Zeming Lin, Natalia Gimelshein, Luca Antiga, et~al.
\newblock Pytorch: An imperative style, high-performance deep learning library.
\newblock {\em Advances in neural information processing systems}, 32, 2019.

\bibitem{peng2021sparse}
Duo Peng, Yinjie Lei, Wen Li, Pingping Zhang, and Yulan Guo.
\newblock Sparse-to-dense feature matching: Intra and inter domain cross-modal
  learning in domain adaptation for 3d semantic segmentation.
\newblock In {\em Proceedings of the IEEE/CVF International Conference on
  Computer Vision}, pages 7108--7117, 2021.

\bibitem{ronneberger2015u}
Olaf Ronneberger, Philipp Fischer, and Thomas Brox.
\newblock U-net: Convolutional networks for biomedical image segmentation.
\newblock In {\em Medical Image Computing and Computer-Assisted
  Intervention--MICCAI 2015: 18th International Conference, Munich, Germany,
  October 5-9, 2015, Proceedings, Part III 18}, pages 234--241. Springer, 2015.

\bibitem{ros2016synthia}
German Ros, Laura Sellart, Joanna Materzynska, David Vazquez, and Antonio~M
  Lopez.
\newblock The synthia dataset: A large collection of synthetic images for
  semantic segmentation of urban scenes.
\newblock In {\em Proceedings of the IEEE Conference on Computer Vision and
  Pattern Recognition}, pages 3234--3243, 2016.

\bibitem{sakaridis2021acdc}
Christos Sakaridis, Dengxin Dai, and Luc Van~Gool.
\newblock Acdc: The adverse conditions dataset with correspondences for
  semantic driving scene understanding.
\newblock In {\em Proceedings of the IEEE/CVF International Conference on
  Computer Vision}, pages 10765--10775, 2021.

\bibitem{shanmugam2021better}
Divya Shanmugam, Davis Blalock, Guha Balakrishnan, and John Guttag.
\newblock Better aggregation in test-time augmentation.
\newblock In {\em Proceedings of the IEEE/CVF International Conference on
  Computer Vision}, pages 1214--1223, 2021.

\bibitem{shin2022mm}
Inkyu Shin, Yi-Hsuan Tsai, Bingbing Zhuang, Samuel Schulter, Buyu Liu, Sparsh
  Garg, In~So Kweon, and Kuk-Jin Yoon.
\newblock Mm-tta: multi-modal test-time adaptation for 3d semantic
  segmentation.
\newblock In {\em Proceedings of the IEEE/CVF Conference on Computer Vision and
  Pattern Recognition}, pages 16928--16937, 2022.

\bibitem{su2022revisiting}
Yongyi Su, Xun Xu, and Kui Jia.
\newblock Revisiting realistic test-time training: Sequential inference and
  adaptation by anchored clustering.
\newblock In {\em Thirty-Sixth Conference on Neural Information Processing
  Systems}, 2022.

\bibitem{sun2020scalability}
Pei Sun, Henrik Kretzschmar, Xerxes Dotiwalla, Aurelien Chouard, Vijaysai
  Patnaik, Paul Tsui, James Guo, Yin Zhou, Yuning Chai, Benjamin Caine, et~al.
\newblock Scalability in perception for autonomous driving: Waymo open dataset.
\newblock In {\em Proceedings of the IEEE/CVF Conference on Computer Vision and
  Pattern Recognition}, pages 2446--2454, 2020.

\bibitem{sun2020test}
Yu Sun, Xiaolong Wang, Zhuang Liu, John Miller, Alexei Efros, and Moritz Hardt.
\newblock Test-time training with self-supervision for generalization under
  distribution shifts.
\newblock In {\em International Conference on Machine Learning}, pages
  9229--9248. PMLR, 2020.

\bibitem{tarvainen2017mean}
Antti Tarvainen and Harri Valpola.
\newblock Mean teachers are better role models: Weight-averaged consistency
  targets improve semi-supervised deep learning results.
\newblock {\em Advances in Neural Information Processing Systems}, 30, 2017.

\bibitem{wang2020tent}
Dequan Wang, Evan Shelhamer, Shaoteng Liu, Bruno Olshausen, and Trevor Darrell.
\newblock Tent: Fully test-time adaptation by entropy minimization.
\newblock In {\em International Conference on Learning Representations}, 2021.

\bibitem{wang2022towards}
Jun-Kun Wang and Andre Wibisono.
\newblock Towards understanding gd with hard and conjugate pseudo-labels for
  test-time adaptation.
\newblock {\em arXiv preprint arXiv:2210.10019}, 2022.

\bibitem{wang2022continual}
Qin Wang, Olga Fink, Luc Van~Gool, and Dengxin Dai.
\newblock Continual test-time domain adaptation.
\newblock In {\em Proceedings of the IEEE/CVF Conference on Computer Vision and
  Pattern Recognition}, pages 7201--7211, 2022.

\bibitem{wang2021multi}
Zejie Wang, Zhen Zhao, Zhao Jin, Zhengping Che, Jian Tang, Chaomin Shen, and
  Yaxin Peng.
\newblock Multi-stage fusion for multi-class 3d lidar detection.
\newblock In {\em 2021 IEEE/CVF International Conference on Computer Vision
  Workshops (ICCVW)}, pages 3113--3121, 2021.

\bibitem{wu2018squeezeseg}
Bichen Wu, Alvin Wan, Xiangyu Yue, and Kurt Keutzer.
\newblock Squeezeseg: Convolutional neural nets with recurrent crf for
  real-time road-object segmentation from 3d lidar point cloud.
\newblock In {\em 2018 IEEE International Conference on Robotics and Automation
  (ICRA)}, pages 1887--1893. IEEE, 2018.

\bibitem{wu2019squeezesegv2}
Bichen Wu, Xuanyu Zhou, Sicheng Zhao, Xiangyu Yue, and Kurt Keutzer.
\newblock Squeezesegv2: Improved model structure and unsupervised domain
  adaptation for road-object segmentation from a lidar point cloud.
\newblock In {\em 2019 International Conference on Robotics and Automation
  (ICRA)}, pages 4376--4382. IEEE, 2019.

\bibitem{xu2021rpvnet}
Jianyun Xu, Ruixiang Zhang, Jian Dou, Yushi Zhu, Jie Sun, and Shiliang Pu.
\newblock Rpvnet: A deep and efficient range-point-voxel fusion network for
  lidar point cloud segmentation.
\newblock In {\em Proceedings of the IEEE/CVF International Conference on
  Computer Vision}, pages 16024--16033, 2021.

\bibitem{xu2022aligning}
Yuecong Xu, Haozhi Cao, Kezhi Mao, Zhenghua Chen, Lihua Xie, and Jianfei Yang.
\newblock Aligning correlation information for domain adaptation in action
  recognition.
\newblock {\em IEEE Transactions on Neural Networks and Learning Systems},
  2022.

\bibitem{xu2022source}
Yuecong Xu, Jianfei Yang, Haozhi Cao, Keyu Wu, Min Wu, and Zhenghua Chen.
\newblock Source-free video domain adaptation by learning temporal consistency
  for action recognition.
\newblock In {\em Computer Vision--ECCV 2022: 17th European Conference, Tel
  Aviv, Israel, October 23--27, 2022, Proceedings, Part XXXIV}, pages 147--164.
  Springer, 2022.

\bibitem{yang2022deep}
Jianfei Yang, Yuecong Xu, Haozhi Cao, Han Zou, and Lihua Xie.
\newblock Deep learning and transfer learning for device-free human activity
  recognition: A survey.
\newblock {\em Journal of Automation and Intelligence}, 1(1):100007, 2022.

\bibitem{Yu_2020_CVPR}
Fisher Yu, Haofeng Chen, Xin Wang, Wenqi Xian, Yingying Chen, Fangchen Liu,
  Vashisht Madhavan, and Trevor Darrell.
\newblock Bdd100k: A diverse driving dataset for heterogeneous multitask
  learning.
\newblock In {\em Proceedings of the IEEE/CVF Conference on Computer Vision and
  Pattern Recognition (CVPR)}, June 2020.

\bibitem{zhang2021memo}
Marvin~Mengxin Zhang, Sergey Levine, and Chelsea Finn.
\newblock Memo: Test time robustness via adaptation and augmentation.
\newblock In {\em NeurIPS 2021 Workshop on Distribution Shifts: Connecting
  Methods and Applications}, 2021.

\bibitem{zhao2023delta}
Bowen Zhao, Chen Chen, and Shu-Tao Xia.
\newblock Delta: degradation-free fully test-time adaptation.
\newblock {\em arXiv preprint arXiv:2301.13018}, 2023.

\bibitem{zou2019confidence}
Yang Zou, Zhiding Yu, Xiaofeng Liu, BVK Kumar, and Jinsong Wang.
\newblock Confidence regularized self-training.
\newblock In {\em Proceedings of the IEEE/CVF International Conference on
  Computer Vision}, pages 5982--5991, 2019.

\end{thebibliography}
}

\end{document}